%% file: main.tex
\documentclass[lettersize,journal]{IEEEtran}
\usepackage{amsmath,amsfonts}
\usepackage{algorithmic}
\usepackage{array}
\usepackage[caption=false,font=normalsize,labelfont=sf,textfont=sf]{subfig}
\usepackage{textcomp}
\usepackage{stfloats}
\usepackage{url}
\usepackage{verbatim}
\usepackage{graphicx}
\def\BibTeX{{\rm B\kern-.05em{\sc i\kern-.025em b}\kern-.08em
    T\kern-.1667em\lower.7ex\hbox{E}\kern-.125emX}}
\usepackage{balance}

\usepackage{booktabs}
\usepackage{multirow}

\usepackage[dvipsnames]{xcolor}

\usepackage[pagebackref,breaklinks,colorlinks]{hyperref}

\usepackage[capitalize]{cleveref}
\crefname{section}{Sec.}{Secs.}
\Crefname{section}{Section}{Sections}
\Crefname{table}{Table}{Tables}
\crefname{table}{Tab.}{Tabs.}

\def\supp{\textit{\textcolor{BrickRed}{supplementary}}}

\begin{document}

\title{On the Robustness of Human-Object Interaction Detection against Distribution Shift}

\author{
Chi Xie,
Shuang Liang,
Jie Li,
Feng Zhu,
Rui Zhao,
Yichen Wei
and~Shengjie Zhao,~\IEEEmembership{Senior Member,~IEEE}
\thanks{This work was supported in part by the National Natural Science Foundation of China under Grant 62076183, in part by the Shanghai Science and Technology Innovation Action Project under Grant 20511100700, in part by the Shanghai Science and Technology Commission Project under Grant 23511103100, in part by the Shanghai Municipal Science and Technology Major Project under Grant 2021SHZDZX0100, and in part by the Fundamental Research Funds for the Central Universities.}
\thanks{Chi Xie, Shuang Liang and Shengjie Zhao are with Tongji University. (email: chixie@tongji.edu.cn, shuangliang@tongji.edu.cn, shengjiezhao@tongji.edu.cn)}

\thanks{Jie Li, Feng Zhu and Rui Zhao are with Sensetime Research. (email: lijie.32@outlook.com, zhufeng@sensetime.com, zhaorui@sensetime.com)}

\thanks{Yichen Wei is with Shukun Technology. (email:wei\_yi\_chen@hotmail.com)}

\thanks{The corresponding author is Shuang Liang.}
}

\markboth{Journal of \LaTeX\ Class Files,~Vol.~18, No.~9, September~2020}%
{How to Use the IEEEtran \LaTeX \ Templates}

\maketitle

\input{sec/0_abstract}


\input{sec/1_intro}
\input{sec/2_related}

\input{sec/3_dataset}
\input{sec/4_analysis}
\input{sec/5_method}
\input{sec/6_experiment}
\input{sec/7_conclusion}


{
\bibliographystyle{IEEEtran}
\bibliography{concise}
}

\input{sec/X_suppl}

\end{document}

%% file: sec/0_abstract.tex
\begin{abstract}
Human-Object Interaction (HOI) detection has seen substantial advances in recent years.
However, existing works focus on the standard setting with ideal images and natural distribution, far from practical scenarios with inevitable distribution shifts.
This hampers the practical applicability of HOI detection.
In this work, we investigate this issue by benchmarking, analyzing, and enhancing the robustness of HOI detection models under various distribution shifts.
We start by proposing a novel automated approach to create the first robustness evaluation benchmark for HOI detection.
Subsequently, we evaluate more than 40 existing HOI detection models on this benchmark, showing their insufficiency, analyzing the features of different frameworks, and discussing how the robustness in HOI is different from other tasks.
With the insights from such analyses, we propose to improve the robustness of HOI detection methods through: (1) a cross-domain data augmentation integrated with mixup, and (2) a feature fusion strategy with frozen vision foundation models.
Both are simple, plug-and-play, and applicable to various methods.
Our experimental results demonstrate that the proposed approach significantly increases the robustness of various methods, with benefits on standard benchmarks, too.
The dataset and code will be released.
\end{abstract}

%% file: sec/1_intro.tex
\section{Introduction}
\label{sec:intro}

\IEEEPARstart{H}{uman}-Object Interaction (HOI) detection~\cite{xu2019interact_hoi_tmm,fang2024HODNtmm,li2025simultaneous_hoi_tmm,xie2025relationlmm} has witnessed significant progress over recent years, showing its potential in numerous applications, including robotics, sports, and healthcare.
Despite these advances, there are still challenges to be addressed for practical applications.

One critical but overlooked limitation of existing works on HOI detection is their unclear effectiveness on inputs with distribution shifts.
Most existing works~\cite{zhang2024hoiplug,li2024hoidiffusion,zhu2025diagnosinghoi} conduct research on a few datasets~\cite{chao2018hoihicodet,gupta2015hoivcoco,liao2020hoippdm} with training and testing data from identical distributions. However, in real-world applications, the models are destined to encounter data with distribution shifts due to environmental changes.
More specifically, existing datasets typically feature images captured under ideal conditions regarding lighting, weather, photographing, etc., which do not represent the diversity of real-world scenarios.
They may fail to generalize well when exposed to complex environments like bad weather, image distortions, or other distribution shift.
\textit{Are the improvements of existing HOI detectors applicable, or just over-fitting to the standard test set? Are they reliable in real-world applications?}
To the best of our knowledge, \textbf{such questions have never been answered, and more importantly, never asked}.

\input{figure/intro_teaser}

\textit{Do we need to specifically investigate distribution shift robustness for HOI detection?}
It is highly likely.
Most existing research on robustness for deep learning focuses on the simple image classification task. There have been diverse robustness benchmarks~\cite{xiao2021robust-imagenet-9-noise-signal,zhang2024robust-imagenet-d-synthetic-object} for image classification, with various types of distribution shifts.
Recently, research on such robustness for object detection~\cite{michaelis2019benchmarking-winter,mao2023robust-coco-o} has grown from simple and naive to diverse and practical.
HOI detection is a more advanced and complicated task. It requires the detection of objects and relationships, and involves multi-modal information like languages and poses. Thus, we cannot know how unique its robustness is, without investigation.

To fill the research gap in HOI detection between limited testbeds and real-world scenarios, we try to comprehensively benchmark, analyze, and enhance the robustness of HOI detection against distribution shift, for the first time.

\input{figure/intro_eval_all}

We start by constructing the first distribution shift robustness benchmark for HOI.
The manual construction~\cite{chao2018hoihicodet,gupta2015hoivcoco,zhuang2018hcvrd} of datasets in multiple domains requires significant human labor costs, and the randomness of manual collection can introduce other variable factors, making it difficult to quantify the impact of distribution shift. Thus, we propose a novel \textbf{AIGC-based domain shift dataset construction approach}. It utilizes image generation models to synthesize image samples under different domain distributions, and avoids the issues of high annotation costs caused by complex triplet objectives and the difficulties of targeted evaluation and analysis. 
Through this, we provide a realistic and comprehensive distribution shift robustness evaluation benchmark for HOI detection.

We conduct extensive evaluations on this new benchmark, assessing the performance of over 40 existing HOI detection methods. The results reveal significant shortcomings in the robustness of these methods, as is shown in~\cref{fig:intro_eval_all}.
We also find that interaction classification, which is a unique sub-task in HOI detection, is the most important source of vulnerability against distribution shift. Through detailed analysis of different components, we identify the characteristics of different HOI detection frameworks facing distribution shifts.
This allows us to draw insights that form the foundation for the following improvements on robustness.

Based on the insights gained from our analyses, we introduce a plug-and-play approach that boosts the robustness of HOI detection methods universally.
It contains two parts.
Firstly, to balance between the robustness over new domains and the performance on the original domain, we devise a \textbf{Cross-Domain Mixup Augmentation} (CMA) method. Its first step creates samples from irrelevant domains in a model-free way, and the second step smoothes the samples from different domains.
Secondly, we notice that existing works utilize CLIP among various choices of Vision Foundation Models (VFMs), and utilize it only to increase the performance on the original domain. Actually, they fail to increase the robustness despite integrating models like CLIP. We propose a \textbf{Feature Fusion with Frozen Foundation Models} (F4M) method, which extracts the local and global information from different VFMs and fuses them into the encoder and decoder of an HOI detection transformer.
Both methods can easily enhance the robustness of various existing HOI detectors.

Our experimental results demonstrate the effectiveness of the proposed approach.
The enhanced models show significant improvements in robustness, outperforming traditional approaches. Additionally, we observe moderate improvements on traditional HOI detection datasets. These indicate our approach not only enhances robustness but also generalizes well across different settings.

In summary, this work tries to address a critical gap in HOI detection by noticing, benchmarking, analyzing, and improving the robustness against distribution shifts in HOI detection for the first time.
The contributions include:
\begin{itemize}
    \item This is the first work to systematically investigate the problem of distribution shift for HOI detection.
    \item We propose an automated approach to construct the first distribution shift robustness benchmark, which enables diverse domains and fine-grained evaluations.
    \item Through extensive experiments on the proposed benchmark, we analyze the robustness problem, find several important insights, and show the direction for exploration.
    \item We propose two solutions to enhance the robustness against distribution shift for various HOI detectors, both of which are conceptually simple, experimentally effective, and universally applicable.
\end{itemize}



%% file: figure/intro_teaser.tex
\begin{figure}
    \centering
    \includegraphics[width=0.9\linewidth]{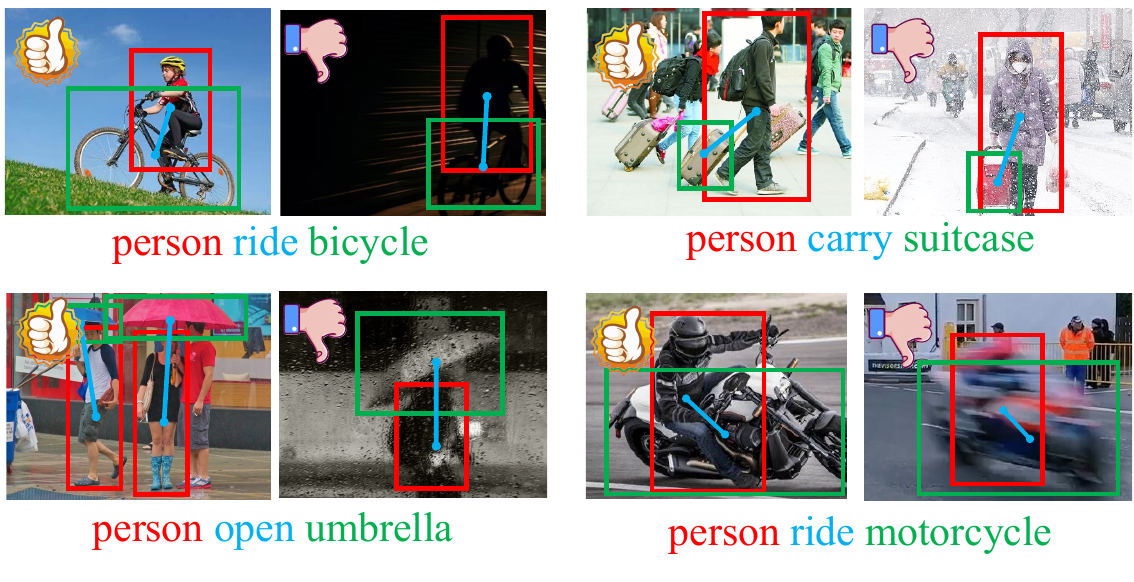}
    \vspace{-4mm}
    \caption{
    \textbf{Existing HOI detection methods focus on natural images (left), on which they achieve significant progress, but ignore situations with distribution shift (right), which are common for in-the-wild evaluation.
    }}
    \vspace{-1em}
    \label{fig:intro_teaser}
\end{figure}

%% file: figure/intro_eval_all.tex
\begin{figure*}
    \centering
    \includegraphics[width=0.95\linewidth]{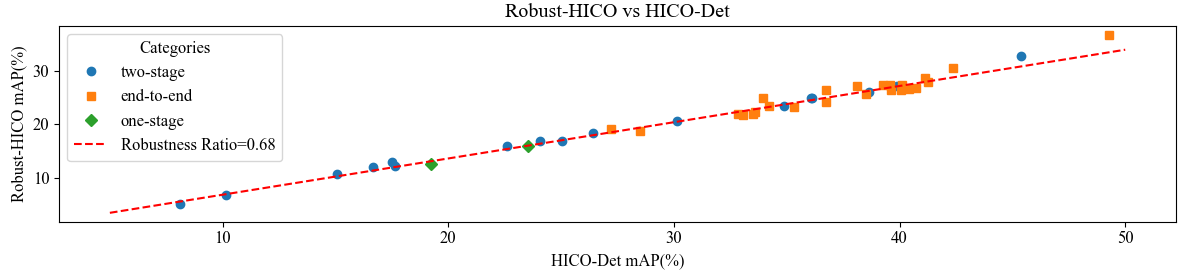}
    \caption{
    \textbf{Existing methods face a significant performance drop on Robust-HICO compared with HICO-DET.}
    They obtain only about 68\% of the performance on the original HICO-DET test set. See~\cref{subsec:analysis_eval} for more.
    }
    \vspace{-1.5em}
    \label{fig:intro_eval_all}
\end{figure*}

%% file: sec/2_related.tex
\section{Related Works}
\label{sec:related}

\subsection{Human-Object Interaction Detection}
\label{subsec:review_hoi}

\noindent \textbf{HOI detection datasets.}
Popular benchmarks include HICO-DET~\cite{chao2018hoihicodet} and V-COCO~\cite{gupta2015hoivcoco}, which cover images under natural distribution.
HOI-A~\cite{liao2020hoippdm} is a dataset focusing on evaluating models under a few real-world application scenarios, but its training and evaluation samples are still from the same distribution.
To the best of our knowledge, there are no benchmarks for investigating distribution shift on HOI detection.

\noindent \textbf{HOI detection methods.}
For this task, early methods adopt a two-stage pipeline~\cite{xu2019interact_hoi_tmm,zhang2021hoispatially}.
Some recent methods~\cite{tamura2021hoiqpic,liao2022hoigen,zhang2023hoipvic,wang2022distance_hoi_acmmm} make use of the powerful DETR~\cite{carion2020end} to build one-stage architecture.
Some of them explore how to mitigate the conflicts between different sub-tasks though disentangled network designs~\cite{xie2023cqlhoi,fang2024HODNtmm}.
Some recent works~\cite{liao2022hoigen,Yuan2023RLIPv2} employ VL models, incorporating knowledge from contrastive~\cite{radford2021CLIP} or generative models~\cite{cao2023UniHOI,xie2025relationlmm}.

Currently, almost all existing works focus on improving HOI detectors on natural images. There are no works investigating the distribution shift, let alone improving it.

\input{figure/dataset_construction}

\subsection{Robustness for Vision Tasks}
\label{subsec:review_robust}

\noindent \textbf{Robustness against distribution shifts.}
Recently, some works investigate the robustness of vision models against distribution shift. The tasks include image recognition~\cite{xiao2021robust-imagenet-9-noise-signal}, object detection~\cite{mao2023robust-coco-o,liu2024benchmarking-object-robust-real-world-ijcv}, etc.
Currently, there is no research investigating the distribution shift robustness in HOI detection, although there are already thousands of methods proposed to improve its performance.

\noindent \textbf{Domain adaptation and generalization.}
The robustness against distribution shift is similar to domain adaptation~\cite{luo2024semi_domain_tmm,wang2022information_domain_tmm} and domain generalization~\cite{jin2021style_domaingeneralization_tmm,niu2023knowledge_domaingeneralization_tmm}.
Domain adaptation focuses on adapting a model trained on a source domain to perform well on a target domain, while domain generalization aims to train a model that performs well on unseen domains without accessing their data.
Some of them~\cite{zhang2022multiple-weather-adaption-pami} also investigate how to make models robust to situations like bad weather.
While such problems have been extensively investigated for tasks like object detection~\cite{pang2024adaptionMCNet,liu2021adaptionDomainContrast}, it remains unexplored for HOI detection.

%% file: figure/dataset_construction.tex
\begin{figure*}
    \centering
    \includegraphics[width=0.75\linewidth]{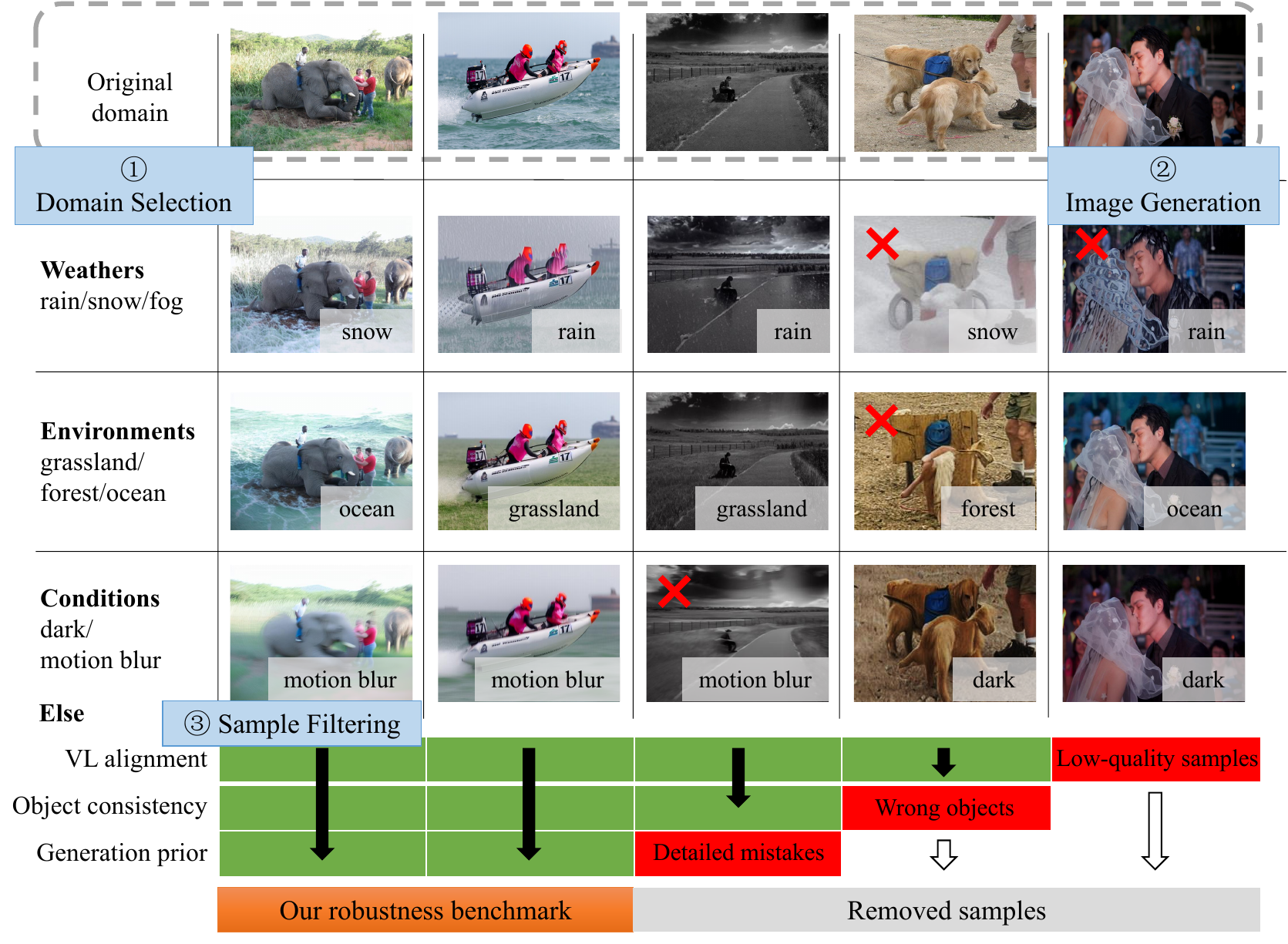}
    \caption{
    \textbf{Overall process of the proposed robustness benchmark construction approach.}
    It mainly contains 3 steps, as elaborated in~\cref{subsec:dataset_construction}.
    }
    \vspace{-1em}
    \label{fig:dataset_construction}
\end{figure*}

%% file: sec/3_dataset.tex
\section{The Robust-HICO Benchmark}
\label{sec:dataset}

We propose an automated approach to construct the first HOI detection benchmark for distribution shift.
We first describe the construction approach in~\cref{subsec:dataset_construction}, then introduce the constructed benchmark in~\cref{subsec:dataset_info}.

\subsection{Automated Construction Approach}
\label{subsec:dataset_construction}

Previous HOI datasets~\cite{chao2018hoihicodet,gupta2015hoivcoco,liao2020hoippdm,zhuang2018hcvrd} were constructed through manual collection and labeling, making them unsuitable for benchmarking distribution shift robustness due to: (1) High cost of multi-domain labeling, and (2) Inability to isolate distribution shift effects.
%
To address this, we develop an automated pipeline with image generation models that generate domain-shifted images while preserving original HOI instances and annotations. 
The potential benefits of such a design are twofold: (1) Data scalability with low cost compared with manual collection and labeling; (2) Fine-grained evaluation and analysis, as the HOI detection labels remain unchanged through different domains and only the domains are changed.
It contains 3 major steps as in~\cref{fig:dataset_construction}:

\noindent \textbf{Domain selection.}
Considering the potential application scenarios, we select 10 domains that should be used in our robustness benchmark from four perspectives with different application challenges and visual difficulties:
(1) Weathers, including \textbf{rain}, \textbf{snow}, and \textbf{fog}. Visual understanding under different weather conditions poses hindrances for recognition, and such robustness is a basic requirement for outdoor scene applications.
(2) Environments, including \textbf{sea}, \textbf{grassland}, and \textbf{forest}. There can be significant changes in the background due to environmental changes, which are key to applying HOI detection in more scenarios.
(3) Imaging conditions, including \textbf{darkness} and \textbf{motion blur}. Under different imaging conditions, the recognition of the entire picture is relatively difficult, posing challenges for the practical application of HOI detection.
(4) Data sources, including \textbf{sketch} and \textbf{watercolor}. In these cases, the appearance features of humans and objects within images are highly abstract, presenting higher challenges and significance in broader application scenarios.

\input{figure/dataset_image_generation}

\noindent \textbf{Domain-guided image generation.}
As illustrated in ~\cref{fig:dataset_image_generation},
based on textual descriptions of each domain, we generate a new image for each image in the test set of HICO-DET~\cite{chao2018hoihicodet}.
We use an advanced text-based image editing model combining null-text inversion~\cite{mokady2023nullTextInversion} and prompt-to-prompt editing~\cite{hertz2022prompt2prompt} for image editing. 
More specifically, we parse the original HOI annotations of each sample with an LLM~\cite{openai2023gpt4}, to obtain the initial prompt. The image is mapped to the visual feature corresponding to this prompt, then, according to the new prompt incorporating domain description, the visual feature is remapped to a new one under the domain, and finally, an edited version of the image is generated based on the new feature. Since this process only changes the image domain without affecting the HOI instances, the annotation for each new sample is the same as the sample in the original domain.

\noindent \textbf{Sample filtering.}
To address potential errors from image editing as shown in~\cref{fig:dataset_sample_filtering}, we filter poor-quality samples using:
(1) \textit{Vision-Language alignment} for filtering low-quality samples, as shown by~\cref{fig:dataset_sample_filtering} (a). 
Since CLIP~\cite{radford2021CLIP} itself has good distribution shift robustness~\cite{crabbe2024interpreting-clip-robust,tu2024closer-look-clip-robust}, we utilize it to decide the quality of images from different distributions and how well it matches the GT annotations,
and discard copies with low relevance across all domains.
(2) \textit{Object consistency} for filtering wrong objects, as shown by~\cref{fig:dataset_sample_filtering} (b). 
We use FIBER~\cite{dou2022fiber}, which achieves leading performance on standard~\cite{lin2014microsoft} and robust object detection~\cite{mao2023robust-coco-o} benchmarks to perform object detection.
If the detection results differ significantly for a copy compared to the original domain, the corresponding copies within all domains are discarded.
(3) \textit{Generation prior} for filtering detailed mistakes, as shown by~\cref{fig:dataset_sample_filtering} (c). 
Images with small objects in GT labels are filtered, as generation models often err on fine details.
Lastly, remaining samples undergo manual review, and poor-quality samples are discarded uniformly across domains.

With the above pipeline, we collected samples from 10 domains for the HICO-DET test set (original domain). For each sample in the original domain, there is a corresponding sample in each domain, sharing the HOI instance annotations.

\input{figure/dataset_sample_filtering}

\input{figure/dataset_vis_sample}

\subsection{Dataset Information}
\label{subsec:dataset_info}

Following the above process, we construct Robust-HICO as an evaluation-only benchmark.
Models should be trained on the original HICO-DET training set, then evaluated on Robust-HICO to assess distribution shift robustness.
\cref{fig:dataset_vis_sample} shows samples from HICO-DET and Robust-HICO. 
They share identical HOI instances but differ in domain characteristics, enabling precise robustness comparisons while controlling for instance variations.

\noindent \textbf{Statistics.}
Robust-HICO has 10 domains with 7,219 images each.
All annotations match the original HICO-DET set.
For fair comparison, we use a subset of HICO-DET (denoted as HICO-DET*) that directly corresponds to Robust-HICO samples.

\noindent \textbf{Characteristics.}
(1) Evaluation only. Shares categories with existing benchmarks~\cite{chao2018hoihicodet} but introduces domain diversity to measure distribution shift robustness.
(2) Annotation consistency. Identical annotations across domains isolate the impact of distribution shifts.
(3) Large scale. Over 70,000 samples across 10 domains, an order of magnitude larger than existing ones~\cite{chao2018hoihicodet,gupta2015hoivcoco,liao2020hoippdm}, enables stable model comparisons.

\subsection{Evaluation metrics}
\label{subsec:dataset_metric}

We assess performance and robustness using three metrics:

\noindent \textbf{Mean Average Precision (mAP).}
The basic metric is HOI detection mAP~\cite{lin2014microsoft,chao2018hoihicodet}, similar to the evaluation on HICO-DET~\cite{chao2018hoihicodet}. By averaging the mAPs for the 10 domains individually, the comprehensive mAP for this dataset is obtained.

\noindent \textbf{Robustness Ratio Margin.}
Inspired by the Effective Robustness metric used in previous works~\cite{hendrycks2021robust-many-faces,mao2023robust-coco-o}, 
we employ a relative metric to exclude the influence of the general capability of models and measure the inherent robustness. 
Specifically, given a method $m$ with its performance $mAP_m^h$ on HICO-DET* and $mAP_m^r$ on Robust-HICO, we calculate
\begin{equation}
RR_m = \frac{mAP_m^r}{mAP_m^h}, \quad
RRM_m = RR_m - \frac{1}{M} \sum_{i=1}^M RR_i, \nonumber
\end{equation}
where $M$ represents the total number of tested methods, the Robust Ratio ($RR$) indicates the percentage of performance on Robust-HICO relative to HICO-DET*, and the Robust Ratio Margin ($RRM$) indicates the advantage of a particular method's robustness over the average of existing methods. Notably, the $RRM$ metric is essentially the same as the effective robustness~\cite{hendrycks2021robust-many-faces,mao2023robust-coco-o} in robust image classification and object detection. 

\noindent \textbf{Fine-grained error analysis.}
The third metric is the robustness related to different sub-tasks of HOI detection.
In the complicated HOI detection task, which part is more vulnerable?
We examine this by calculating the percentages of different error types for the sub-tasks.
Then, we compare the error percentages on the original and the new domains.


%% file: figure/dataset_image_generation.tex
\begin{figure}
    \centering
    \includegraphics[width=\linewidth]{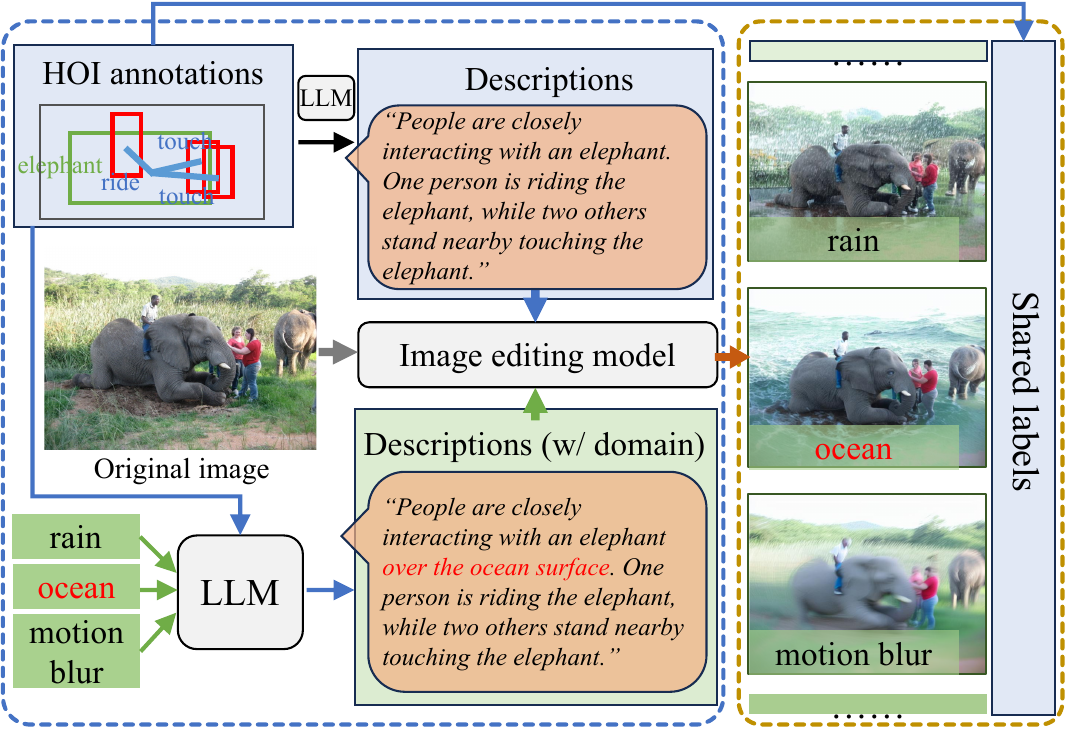}
    \caption{
    \textbf{The domain-guided image generation step in the proposed benchmark construction pipeline.}
    An image editing model accepts the image descriptions (w/o and w/ domain information) as well as the original image and generates an image for the new domain.
    It is described in~\cref{subsec:dataset_construction}.
    }
    \vspace{-6pt}
    \label{fig:dataset_image_generation}
\end{figure}

%% file: figure/dataset_sample_filtering.tex
\begin{figure}
    \centering
    \includegraphics[width=0.8\linewidth]{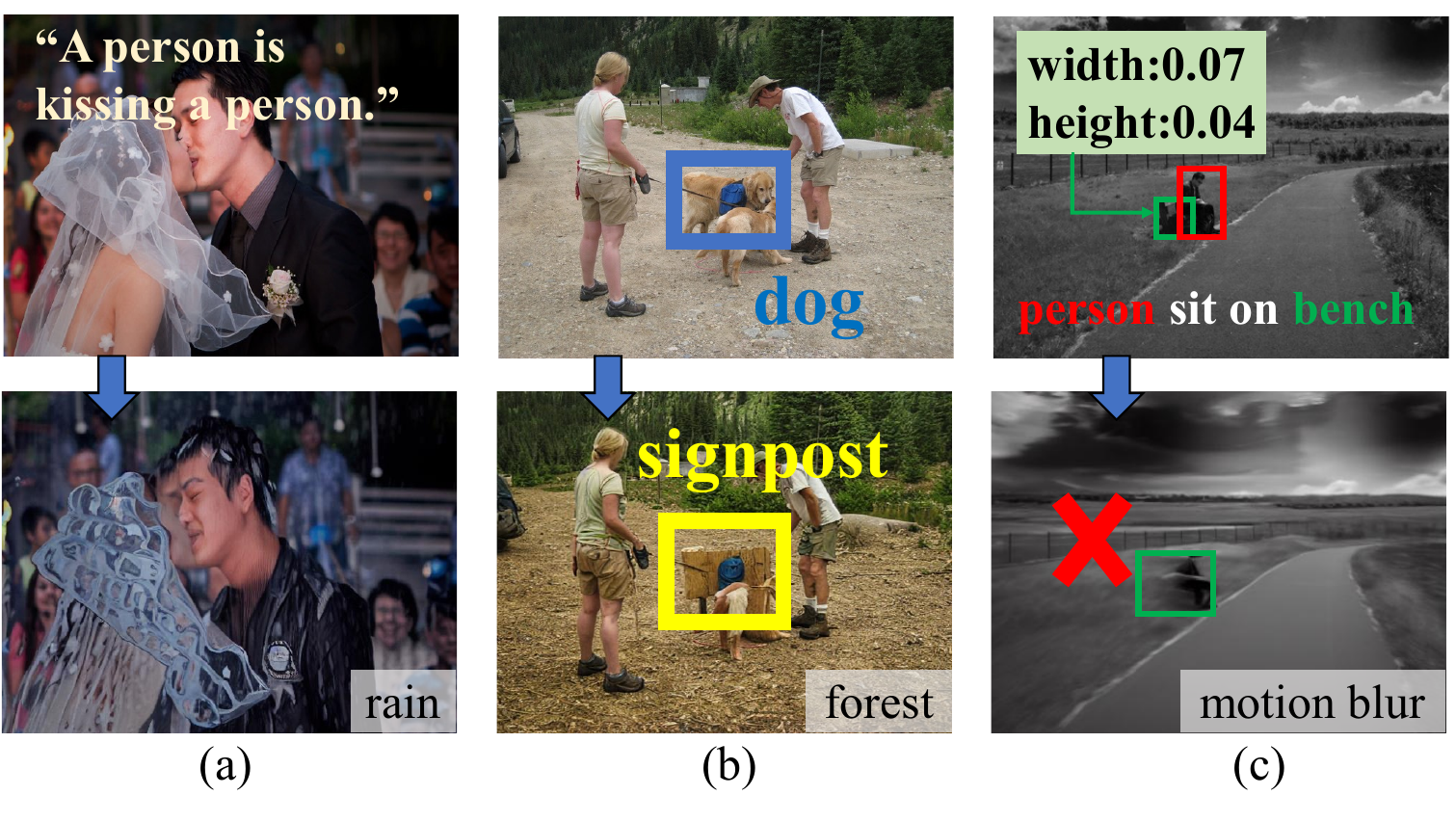}
    \caption{
    \textbf{Examples of filtered samples in the 3 sample filtering strategies in our benchmark construction pipeline.}
    (a) Filtering via VL alignment; (2) filtering with object consistency; (3) filtering with generation prior. Elaborated in~\cref{subsec:dataset_construction}.
    }
    \vspace{-10pt}
    \label{fig:dataset_sample_filtering}
\end{figure}

%% file: figure/dataset_vis_sample.tex
\begin{figure*}
    \centering
    \includegraphics[width=0.9\linewidth]{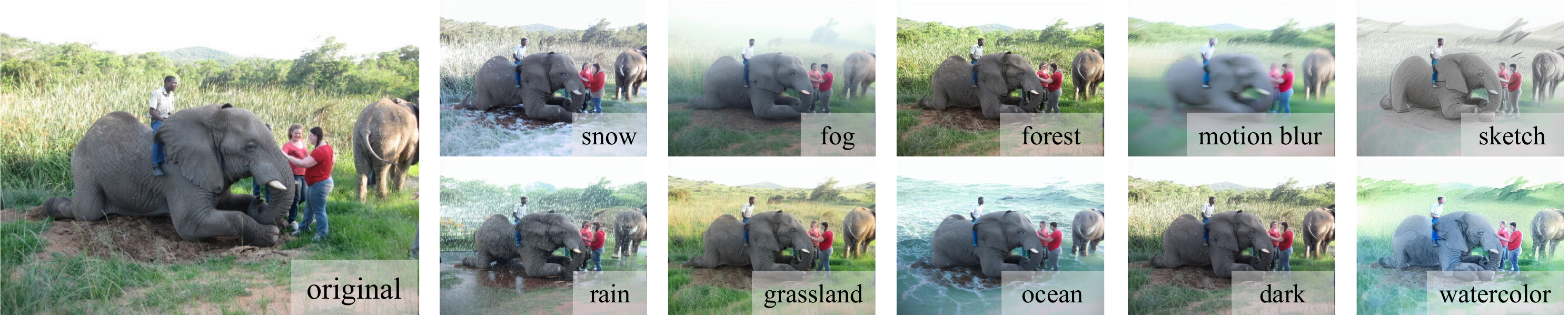}
    \caption{
    \textbf{Some samples from the proposed Robust-HICO benchmark.}
    Each sample in the original domain corresponds to 10 samples from other domains, all sharing the same HOI instances and annotations. Discussed in~\cref{subsec:dataset_info}.
}
    \label{fig:dataset_vis_sample}
\end{figure*}

%% file: sec/4_analysis.tex

\section{Analysis of Existing Methods}
\label{sec:analysis}


We evaluate existing methods on Robust-HICO in \cref{subsec:analysis_eval}, and examine the vulnerability against distribution shift for the sub-tasks in \cref{subsec:analysis_subtasks},
We examine factors that may affect such robustness in \cref{subsec:analysis_components} and draw some insights for stronger robustness.

\subsection{Comprehensive Evaluation of Existing Works}
\label{subsec:analysis_eval}


We conduct a systematic evaluation of existing methods' robustness using Robust-HICO. 
As illustrated in \cref{fig:intro_eval_all}, while recent methods demonstrate substantial performance gains on HICO-DET*, their performance on Robust-HICO remains at approximately 70\% of their HICO-DET* scores. This linear correlation suggests that performance improvements primarily reflect enhanced general capabilities rather than improved robustness against distribution shifts. 
With the proposed metrics, we calculate an average Robustness Ratio ($RR$) of 0.68 across existing methods.

\subsection{Breakdown over Sub-tasks}
\label{subsec:analysis_subtasks}


\input{figure/analysis_error}

Our fine-grained error analysis (\cref{subsec:dataset_metric}) reveals which HOI detection components are most susceptible to distribution shifts. 
\cref{fig:analysis_error} presents error type distributions for QPIC~\cite{tamura2021hoiqpic} across HICO-DET* and Robust-HICO. The results demonstrate:
1) A significant increase in interaction classification errors under distribution shifts.
2) A moderate rise in object classification errors.
3) Relative stability in other error types.
These findings establish interaction classification as the most distribution-sensitive sub-task in HOI detection. Notably, this vulnerability represents a previously unaddressed challenge in robustness research, distinct from known issues in other vision tasks. The results strongly validate the importance of our benchmark for advancing HOI detection robustness.

\subsection{Breakdown over Method Components}
\label{subsec:analysis_components}

Through controlled experiments, we investigate how the differences in each model components can affect robustness against distribution shift. Due to the page limit, they are presented in \supp{} Sec. I.

In summary, we discover:
(1) Current methods exhibit limited distribution shift robustness.
(2) End-to-end transformers improve performance but may compromise robustness.
(3) More pre-training data helps improve distribution shift robustness.
(4) Existing VFM integrations do not take advantage of the robustness of the foundation model.
The surprising disconnect between VFM robustness and HOI performance suggests the need for better adaptation methods to transfer foundation model capabilities.

These analyses yield five key insights for developing more robust HOI detection models:
(1) \textbf{Backbone network.}
While backbone networks significantly impact robustness, direct modification is computationally expensive. Our proposed F4M approach (\cref{sec:method}) offers a practical alternative by effectively integrating external backbone networks.
%
(2) \textbf{Scaling data.}
Larger-scale training data benefits the model. We explore this direction in the proposed CMA in \cref{sec:method}.
(3) \textbf{Architecture trade-offs.}
The well-recognized end-to-end frameworks based on DETR have advantages in structural simplicity and strong performance, but do not show superiority in robustness to distribution shift.
(4) \textbf{VFMs utilization challenge.}
%
Despite CLIP's inherent robustness~\cite{radford2021CLIP,crabbe2024interpreting-clip-robust,tu2024closer-look-clip-robust}, current CLIP-based HOI detectors fail to transfer this robustness. We attribute this to the known robustness degradation from fine-tuning~\cite{radford2021CLIP}.
(5) \textbf{Object detector limitations.}
Improvements to object detectors themselves have little impact on robustness and should not be the focus of this work.

%% file: figure/analysis_error.tex
\begin{figure*}
    \centering
    \includegraphics[width=0.8\linewidth]{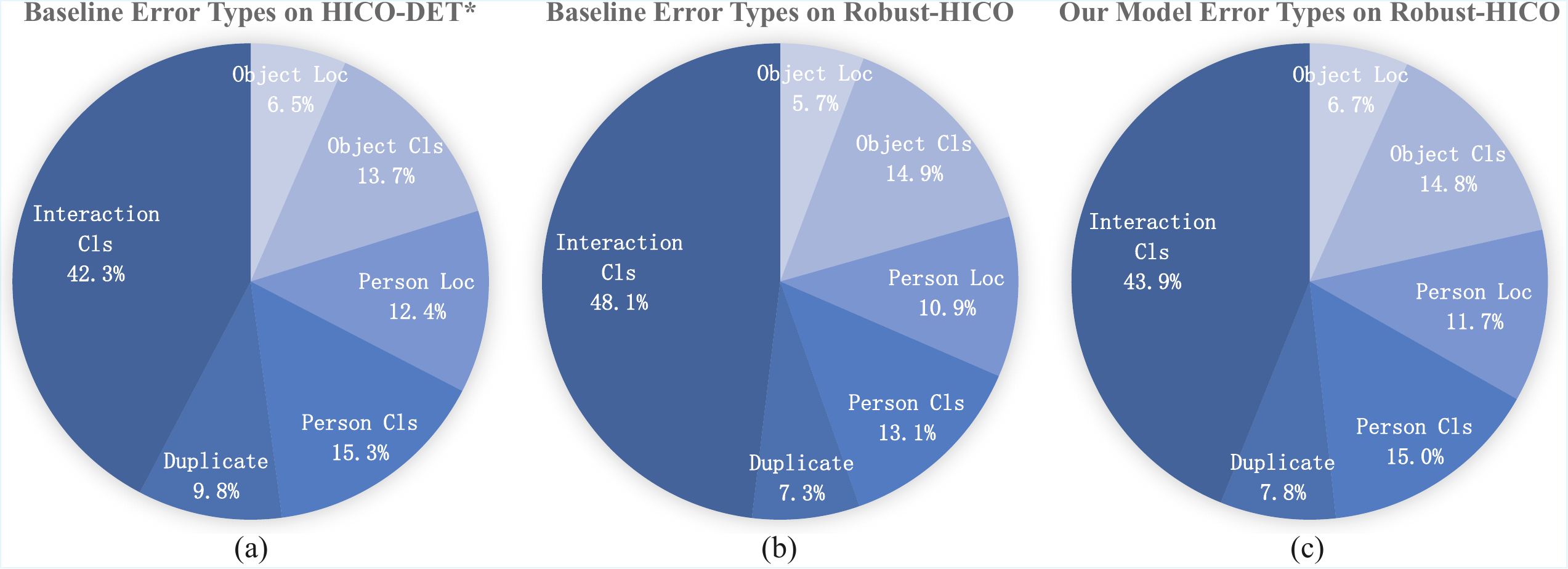}
    \caption{
    \textbf{Analysis of error types.}
    (a) The baseline~\cite{tamura2021hoiqpic} on the original domain (HICO-DET*~\cite{chao2018hoihicodet}); (b) The baseline~\cite{tamura2021hoiqpic} on the domains with distribution shifts (Robust-HICO); (c) Our models improved with the methods in~\cref{sec:method} on Robust-HICO.
    See~\cref{subsec:analysis_subtasks} for discussion.
    }
    \vspace{-1em}
    \label{fig:analysis_error}
\end{figure*}

%% file: sec/5_method.tex
\section{Our Approach}
\label{sec:method}

\input{figure/method_cma}
As discussed in \cref{subsec:analysis_eval}, the robustness to distributional shift in HOI detection has not been significantly improved in recent works. In this section, we attempt to enhance the robustness against distributional shift in HOI detection.

\subsection{Guidelines for Methodological Design}
\label{subsec:method_pre}



Building upon our robustness analysis in \cref{subsec:analysis_components}, we establish four key design principles for developing effective robustness enhancement methods:
(1) \textbf{Model-agnostic framework.}
Given the diversity of existing architectures and their shared robustness limitations, our approach should be applicable across different model.
(2) \textbf{Data integrity preservation.}
%
While data augmentation through generative models could theoretically help, we avoid this approach to prevent potential data leakage between our evaluation benchmark (constructed using image editing models) and training data.
(3) \textbf{Performance preservation.}
Do not harm the performance on the original domain. Stronger robustness at the cost of the original performance may not be suitable for applications.
(4) \textbf{Pretraining utilization.}
%
The method should effectively leverage existing large-scale pretraining (\emph{e.g.}, on ImageNet~\cite{krizhevsky2012imagenet}, COCO~\cite{lin2014microsoft}, or CLIP~\cite{radford2021CLIP}) without requiring retraining from scratch.

Based on the insights in \cref{subsec:analysis_components} and these guidelines, in \cref{subsec:method_cma} we propose a CMA method that improves HOI training data. It satisfies all 4 guidelines. In \cref{subsec:method_f4m} we propose a F4M method that utilizes frozen VFMs. It satisfies guidelines (2) (3) and (4) and only requires the model framework to be transformer-based, which is practically OK for guideline (1).

\input{figure/method_domains}

\subsection{Cross-Domain Mixup Augmentation}
\label{subsec:method_cma}


Our analysis in \cref{subsec:analysis_components} reveals that training data quantity and diversity are crucial for improving HOI detector robustness against distribution shifts. 
However, acquiring such diverse data is often cost-prohibitive. 
Also,
based on the guidelines in \cref{subsec:method_pre}, simply generating more training samples with generation models could violate guidelines (2) (3) and (4) above.
Thus, we attempt to obtain multi-domain data based on existing samples, and combine it with data augmentation techniques to improve the robustness across different domains.

In this part, CMA first enhances the samples with richer domain characteristics in a parameter-free way, then combines and improves both mixup~\cite{zhang2018mixup} and dropout~\cite{srivastava2014dropout,gal2017concreteDropout,xie2023temporaldropout} to obtain smoother samples beneficial for both new and original domains. As shown in \cref{fig:method_cma}, it includes two steps:

\noindent \textbf{Sample synthesis.}
In this step, we create new samples based on existing samples.
To avoid potential leaks of training data, we create samples through parameter-free image editing operations, rather than generative image editing models.
In implementation, we utilize an image processing toolkit provided by previous work~\cite{michaelis2019benchmarking-winter}.
To make CMA suitable for more than the robust-HICO benchmark, we use more common domain types, namely image corruptions~\cite{michaelis2019benchmarking-winter}. We exclude domains that might overlap with Robust-HICO, to evaluate the effects under distribution shift more accurately. The domains include 12 types presented in ~\cref{fig:method_domains}.
Notably, this editing process relies on simple image processing functions without using image editing or generation models.

After obtaining the above samples, a direct way to use them is to replace the corresponding samples in the original training set or add them directly to the training set. However, this may harm the performance on the original domain. To address this issue, we further adopt a second step, sample mixing.

\noindent \textbf{Sample mixing.}
To better leverage the newly synthesized samples and to balance the generalization on the new distributions with the performance on the original distribution, we further combine and improve upon the mixup~\cite{zhang2018mixup}.

Mixup combines two sample images and their labels through linear interpolation to create a new mixed sample, thereby enhancing model performance and generalization. To our knowledge, it has not yet been applied to HOI detection. Here we design a mixing step specifically for HOI detection and apply it to the synthesized samples above.

Given two images $I_A$ and $I_B$ and their corresponding annotations $Y_A$ and $Y_B$, we apply spatial dropout to one of the images, say $I_A$. Specifically, for patches not covered by object bounding boxes, by a probability $\pi_{c}$, we set all the pixel values in each patch to 0.

Then, a mixing ratio $\mu$ is drawn from a beta distribution, i.e., $\mu \sim Beta(\alpha,\alpha)$, where $\alpha$ is a hyperparameter of mixup. This is followed by calculating
$$
I_{mix} = \mu * I_A + (1 - \mu) * I_B,
$$
where $I_{mix}$ is the mixed image. For the annotations $Y_A$ and $Y_B$ from the two images, we directly merge them. Through this process, any two samples are mixed into a new sample.

During training, we apply sample mixing among the original and the synthesized samples, as well as among samples from different synthetic domains. By combining sample synthesis and sample mixing, CMA effectively enhances the cross-domain robustness of various baseline models and also benefits standard HOI detection performance.

\input{figure/method_f4m}

\subsection{Fusion with Frozen Foundation Models}
\label{subsec:method_f4m}

According to the analysis in \cref{subsec:analysis_components}, stronger backbones are beneficial to the robustness, so it is natural to utilize strong VFMs. Some like CLIP~\cite{radford2021CLIP} inherently possess good robustness to distribution shift, but existing HOI detectors utilizing them failed to inherit such robustness.

Inspired by some recent works~\cite{fu2024frozendetr,xie2023cqlhoi}, we propose F4M, which utilizes frozen VFMs as plug-and-play feature enhancer.
Firstly, by integrating global tokens from the foundation model, F4M offers compact representations of different domain scenarios, aiding the decoding of instance queries in the decoder.
Additionally, by integrating regional tokens from the foundation model, F4M provides semantic details to enrich the features in the encoder. The overall process of F4M is illustrated in \cref{fig:method_f4m}.

\noindent \textbf{Global token enhanced decoder.}
Inspired by previous works~\cite{xie2023cqlhoi,fu2024frozendetr}, we introduces image-level queries into HOI decoders. Unlike traditional instance queries, image queries represent the entire image. As VFMs have strong capabilities for extracting robust global feature, we leverage these capabilities by treating the global token from a VFM as an image-level query. With this robust query as context, the decoder can predict results more stably across different domains.

Specifically, as shown in~\cite{fu2024frozendetr}, for each image, the frozen VFM extracts the image query.
The image is resized to match the input size of the foundation model. Then, in each decoder layer, the image query is projected into the same dimension as the instance queries, and these two types of queries are concatenated before being fed into the self-attention module. In the subsequent self-attention, instance queries can adaptively interact with the image query to extract high-level image understanding information from the foundation model. Finally, the image query is discarded after passing through the self-attention module.
To further preserve fine-grained details in the global token, we adopt a lightweight design following previous works utilizing CLIP~\cite{fu2024frozendetr,chen2024internvl2}, which is elaborated in \supp{} Sec. II.

Ultimately, all image queries are concatenated with the instance queries into a sequence. When multiple VFMs are used simultaneously, the image queries from different models are also concatenated into a sequence.

\noindent \textbf{Regional token enhanced encoder.}
We further utilize the regional tokens with fine-grained semantic details. Specifically, as shown in~\cref{fig:method_f4m}, regional tokens are obtained from the last layer of the foundation model and reshaped into 2D feature maps. Then, these 2D feature maps are concatenated with the feature maps from the backbone and passed to the encoder, achieving adaptive fusion using the existing encoder structure. After the fusion by the encoder, the updated backbone features absorb advanced semantic understanding information from the VFM, which is helpful for different scenarios. Notably, when multiple VFMs are used simultaneously, the feature maps of regional tokens from different models are also concatenated.

Additionally, to avoid imposing excessive memory and speed burdens on the Transformer encoder, and prevent overreliance on certain regions of a specific foundation model, spatial dropout~\cite{srivastava2014dropout,gal2017concreteDropout} is applied to the regional token features of the  before they are input into the decoder. Specifically, during training, regional tokens from each VFM are randomly dropped based on a probability $\pi_{f}$, and the remaining regional token features are scaled up by the ratio of the original number of regional tokens to the remaining ones; during testing, no regional tokens are dropped, and no scaling is necessary.


In this part, F4M uses the regional tokens extracted by the VFM to enhance the encoder of the HOI detection model, and utilizes the global tokens to improve the decoding process.
It significantly enhances the robustness to distributional shift compared to previous works utilizing VFMs. It also prevents issues caused by architectural differences between the original backbone and the foundation model.


%% file: figure/method_cma.tex
\begin{figure}
    \centering
    \includegraphics[width=0.85\linewidth]{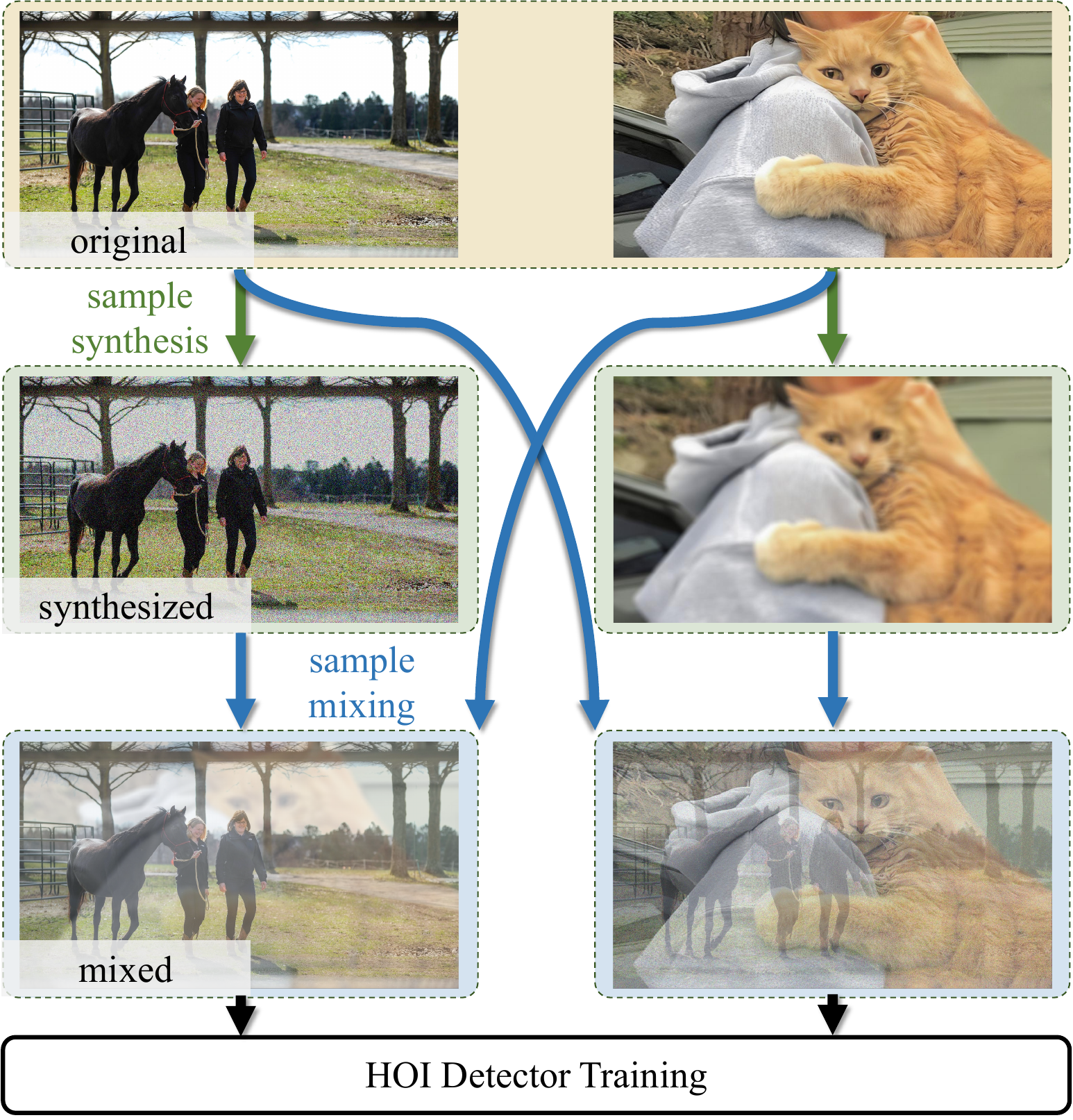}
    \caption{
    \textbf{Overview of the proposed CMA.}
    Starting from original samples~\cite{chao2018hoihicodet}, the sample synthesis and sample mixing steps generate the final samples. Elaborated in~\cref{subsec:method_cma}.
    }
    \vspace{-10pt}
    \label{fig:method_cma}
\end{figure}

%% file: figure/method_domains.tex
\begin{figure}
    \centering
    \includegraphics[width=\linewidth]{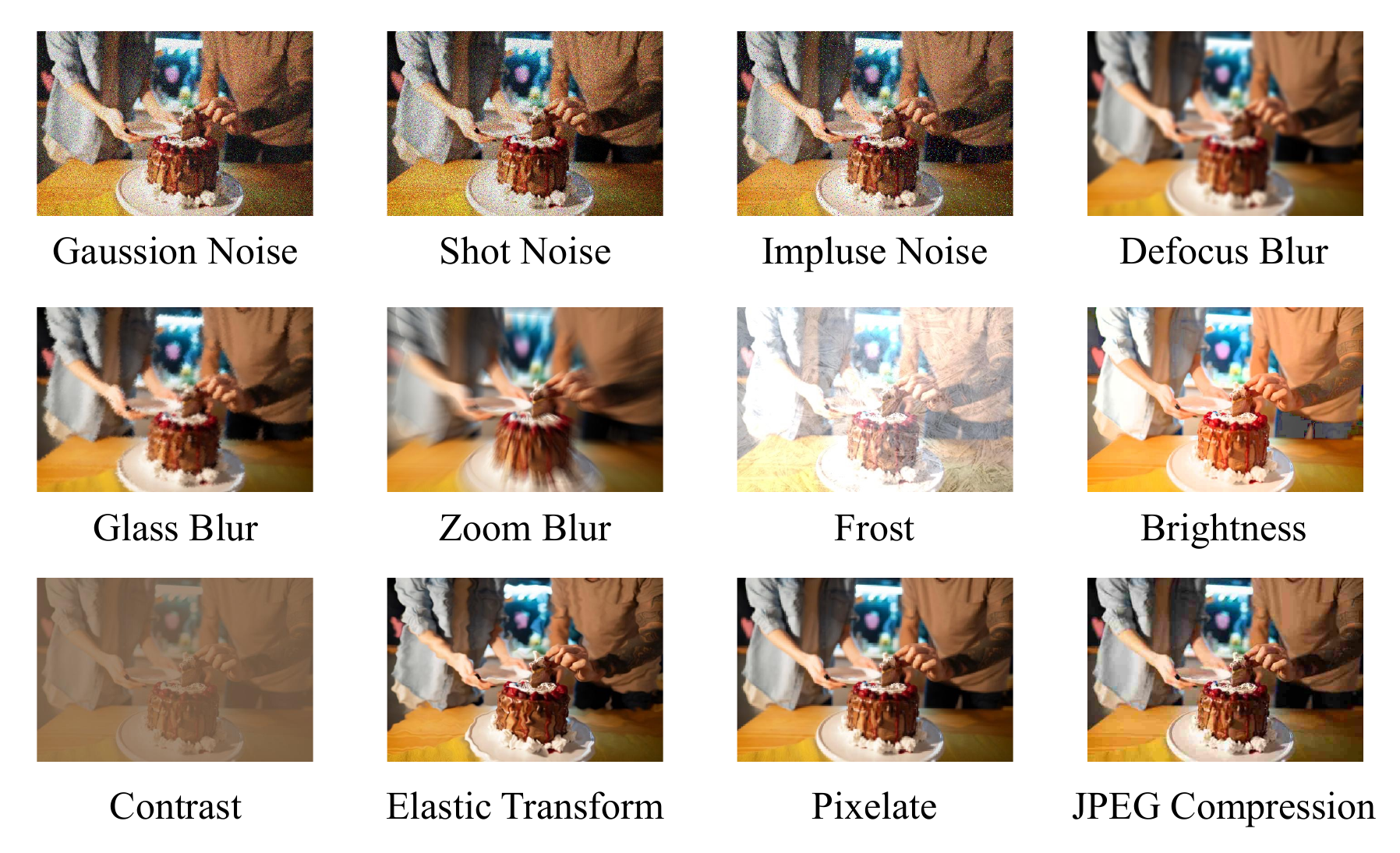}
    \vspace{-2em}
    \caption{
    \textbf{Domains involved in the proposed CMA.}
    These domains do not overlap with the domains in Robust-HICO.
    Elaborated in~\cref{subsec:method_cma}.
    }
    \vspace{-1.5em}
    \label{fig:method_domains}
\end{figure}

%% file: figure/method_f4m.tex
\begin{figure}
    \centering
    \includegraphics[width=0.9\linewidth]{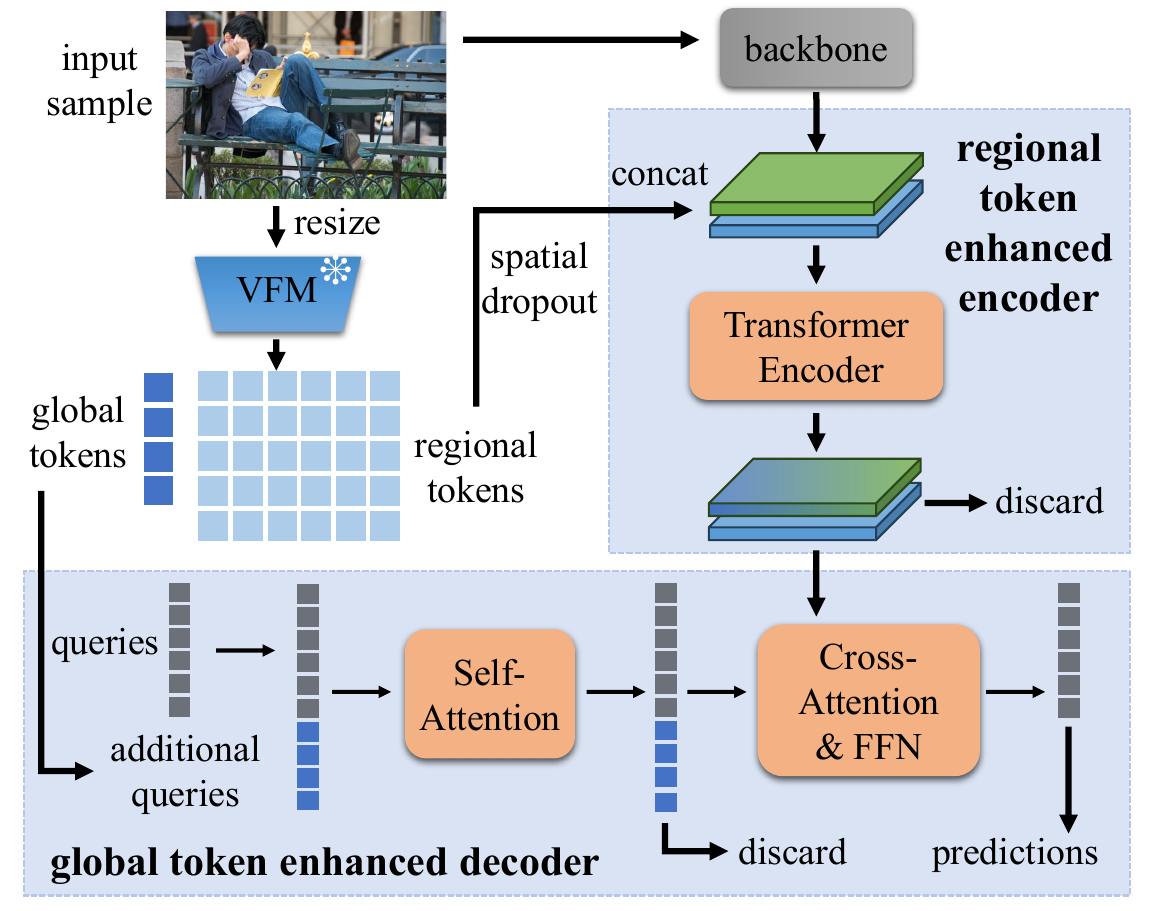}
    \caption{
    \textbf{Overview of the proposed F4M method.}
    It fuses the feature from VFMs into the encoder and decoder of HOI detectors.
    Elaborated in~\cref{subsec:method_f4m}.
    }
    \vspace{-1.5em}
    \label{fig:method_f4m}
\end{figure}

%% file: sec/6_experiment.tex
\section{Experiments for the Proposed Method}
\label{sec:experiments}


\input{table/exp_sota}
\subsection{Experimental Setup}
\label{subsec:exp_setup}

\noindent \textbf{Standard HOI detection.}
Following most previous works~\cite{tamura2021hoiqpic,xie2023cqlhoi,liao2022hoigen}, we adopt the widely-used HICO-DET~\cite{chao2018hoihicodet} and V-COCO~\cite{gupta2015hoivcoco} separately, both containing a training set and a test set.
Due to the page limit, we leave the details in~\supp{} Sec. III.

\noindent \textbf{HOI robustness.}
As introduced in \cref{sec:analysis}, we use the HICO-DET~\cite{chao2018hoihicodet} training set for training and Robust-HICO for evaluation.
When comparing the same method on HICO-DET and Robust-HICO, to ensure strict comparability of results, we limit the evaluation on HICO-DET to a subset that has corresponding cross-domain replicas within Robust-HICO, which is marked as HICO-DET*.

The evaluation metrics on both HICO-DET and Robust-HICO are mAP. However, as stated in \cref{sec:analysis}, when analyzing and comparing the distribution shift robustness of models on Robust-HICO, we adopt RRM, which reflects the relative distribution shift robustness of HOI detection models.

\noindent \textbf{Implementation details.}
Since F4M and CMA are plug-and-play and adaptable to different baseline models, the hyperparameters for models integrating them are consistent with their baselines.
For F4M, we use two VFMs by default, CLIP~\cite{radford2021CLIP} and DINOv2~\cite{oquab2023dinov2}, both based on ViT-L~\cite{dosovitskiy2021ViT}.
The input size for the VFMs is 336.
For CMA, the mixing ratio sampling distribution hyperparameter $\alpha$ is set to 1.5 by default.
The spatial dropout probabilities $\pi_{f}$ and $\pi_{c}$ in F4M and CMA are set to 0.5 and 0.3, respectively.

\input{table/exp_robust}

\subsection{Comparison with Existing Works}
\label{subsec:exp_sota}

Through experiments on HICO-DET and V-COCO, we first demonstrate that the proposed approach is beneficial to the performance of standard HOI detection, then verify how it boosts distribution shift robustness on Robust-HICO.

\noindent \textbf{Standard HOI detection.}
As seen from~\cref{tab:exp_sota}, the proposed F4M and CMA improve the performance of HOI detection methods.
For example, F4M boosts the performance of PViC~\cite{zhang2023hoipvic} on HICO-DET and V-COCO by 1.6 mAP and 1.5 AP separately. CMA also improves PViC. 
Furthermore, when F4M and CMA are applied together, they bring the largest performance gain.
Besides, we can observe that the proposed two methods, especially F4M, significantly boost performance on rare categories, which is desirable.

\noindent \textbf{Distribution shift robustness.}
As shown in~\cref{tab:exp_robust}, after integrating F4M and CMA, each model exhibits stronger distribution shift robustness relative to its baseline. Notably, for multiple baselines, the improvements in robustness by these two methods are significant, and the improvement is even more pronounced when both are used together. This shows CMA and F4M are experimentally complementary to each other.
The reason may be that F4M inherits excellent domain generalization ability from the VFM, while CMA enables the model to learn features that are less dependent on domain distribution and thus more generalized at the data level.

\subsection{Ablation Study}
\label{subsec:exp_ablation}

We perform ablation studies on the proposed F4M and CMA. These experiments are based on the performance on HICO-DET* and Robust-HICO, and the corresponding distribution shift robustness metric RRM. By default, we adopt PViC~\cite{zhang2023hoipvic} as the baseline.

\input{table/exp_dropout}

\noindent \textbf{Spatial dropout in CMA and F4M.}
In F4M, spatial dropout is applied to the regional token feature maps of the VFMs to prevent the HOI detection model from overly relying on specific region features of a particular foundation model. In CMA, spatial dropout is applied during the interpolation process of input images to avoid the model becoming overly dependent on some part of the image background that might be domain-specific.
As shown in~\cref{tab:exp_dropout}, they both have a certain impact on the results. When both are set to 0, which means that no spatial dropout is used, the model still has decent robustness, but applying some spatial dropout leads to more significant improvements in both performance and robustness.

\input{table/exp_mixup}
\input{table/exp_domains}

\subsubsection{Ablation study for CMA}

\noindent \textbf{Method components.}
From the experimental results in~\cref{tab:exp_mixup}, it can be seen that although the sample synthesis step can enhance the robustness metric RRM, it directly causes a noticeable performance drop on the original domain. Besides, the sample mixing step alone only slightly improves performance on the original domain, and has a relatively weak effect on robustness. By combining the two, CMA achieves not only a performance boost on the original domain but also elevates the robustness to a new level.

\noindent \textbf{Domain diversity.}
As seen in~\cref{tab:exp_domains}, overall, the richer the training domains are used in CMA, the better the performance and robustness will be. Although the standard performance on the original domain with 9 domains slightly exceeds that with 12 domains, the difference is negligible. Even with as few as 3 domains, CMA still has a certain positive effect on both the performance and the robustness, demonstrating its scalability.

\input{table/exp_f4m}

\subsubsection{Ablation study for F4M}

\noindent \textbf{Method components.}
\cref{tab:exp_f4m} presents ablation studies on different components of F4M. The designs of enhancing the decoder with global tokens and the encoder with regional tokens are both verified to be useful. However, enhancing the encoder with global tokens or the decoder with regional tokens brings much less improvement. This validates the rationality of the method design.
Moreover, if F4M is not adopted and the VFM is used merely as the backbone for HOI detection, there is a significant drop in model performance, and the robustness metric RRM increases slightly.
The performance drop may be due to the small input size of the VFM, which could lead to performance loss when used as the backbone; the small increase in robustness aligns with many existing works utilizing VFMs and highlights the necessity and value of F4M.

We perform more ablation studies in \supp{} Sec. IV, including the influence of VFMs, the importance of freezing the VFM, and the effectiveness of the query design.

\input{figure/exp_qualitative}

\subsection{More Analyses}
\label{subsec:exp_analysis}

\noindent \textbf{Fine-grained error analysis.}
In~\cref{fig:analysis_error}, we compare the error types between a baseline~\cite{tamura2021hoiqpic} and the improved model on the Robust-HICO dataset. Our method relatively reduces the error ratios in interaction classification and object classification. This indicates the primary way the proposed approach enhances robustness to distributional shift is by improving classification sub-tasks, particularly interaction classification.

\noindent \textbf{Qualitative analysis.}
In~\cref{fig:exp_qualitative}, we compare the visualization results of the baseline~\cite{tamura2021hoiqpic} and the proposed model. Clearly, for samples with the same interaction instances across different domains, the baseline method exhibits significant interference due to domain changes; in contrast, the proposed model yields results across different domains that closely resemble those from the original domain, maintaining stability and reliability.
This qualitatively demonstrates the effectiveness of our approach in terms of robustness to distributional shift.
Additional analyses on the error cases over both the original and new domains are kept in~\supp{} Sec. IV.


%% file: table/exp_sota.tex
\begin{table}[t]
\centering

\caption{
Comparison on the popular HICO-DET and V-COCO benchmarks, between previous works and those improved by the proposed F4M and CMA.
}
\vspace{-6pt}

\resizebox{0.8\linewidth}{!}{
    \begin{tabular}{l|ccc|cc}
        \hline
        Method & \multicolumn{3}{c}{HICO-DET} & \multicolumn{2}{c}{V-COCO} \\
        \cline{2-6}
        & Full & Rare & Non-Rare & S\#1 & S\#2 \\
        \midrule
        CQL~\cite{xie2023cqlhoi} & 36.1 & 34.1 & 36.7 & 64.8 & 69.5 \\
        RLIPv2~\cite{Yuan2023RLIPv2} & 35.4 & 29.6 & 37.1 & 65.9 & 68.0 \\
        Pose-Aware~\cite{wu2024hoiposehybrid} & 35.9 & 32.5 & 36.9 & 61.1 & 66.6 \\
        MP-HOI~\cite{yang2024hoiopen} & 36.5 & 35.5 & 36.8 & 66.2 & 67.6 \\
        
        \hline
        QPIC~\cite{tamura2021hoiqpic} & 28.9 & 21.6 & 31.1 & 61.4 & 63.7 \\
        +CMA & 30.4 & 23.7 & 32.4 & 62.7 & 64.8 \\
        +F4M & 31.1 & 27.5 & 32.2 & 63.1 & 64.9 \\
        +F4M+CMA & 32.0 & 27.9 & 33.2 & 63.8 & 65.6 \\
        \hline
        GEN-VLKT~\cite{liao2022hoigen} & 33.7 & 29.9 & 34.8 & 64.9 & 66.7 \\
        +CMA & 34.6 & 29.8 & 36.0 & 65.6 & 67.1 \\
        +F4M & 34.8 & 32.6 & 35.5 & 66.0 & 67.5 \\
        +F4M+CMA & 35.0 & 30.7 & 36.3 & 66.3 & 67.5 \\
        \hline
        PViC~\cite{zhang2023hoipvic} & 34.7 & 32.1 & 35.5 & 62.8 & 67.8 \\
        +CMA & 35.1 & 33.0 & 35.7 & 63.0 & 68.3 \\
        +F4M & 36.3 & 35.4 & 36.6 & 64.3 & 69.2 \\
        +F4M+CMA & 36.6 & 35.3 & 37.0 & 64.7 & 68.6 \\
        \hline
    \end{tabular}}

\vspace{-6pt}
\label{tab:exp_sota}
\end{table}

%% file: table/exp_robust.tex
\begin{table}[t]
\centering

\caption{Comparison with previous methods on robustness.
}
\vspace{-6pt}

\resizebox{0.7\linewidth}{!}{
\begin{tabular}{l|c|c}
\toprule
Method & HICO-DET* & RRM (\%) \\
\midrule
QPIC~\cite{tamura2021hoiqpic} & 33.64 & -1.7 \\
+CMA & 34.58 & 1.6 \\
+F4M & 34.97 & 2.5 \\
+F4M+CMA & \textbf{35.71} & \textbf{4.1} \\
\hline
GEN-VLKT~\cite{liao2022hoigen} & 38.50 & -1.4 \\
+CMA & 38.19 & 1.9 \\
+F4M & 40.74 & 2.7 \\
+F4M+CMA & \textbf{40.97} & \textbf{3.6} \\
\hline
PViC~\cite{zhang2023hoipvic} & 38.65 & -0.7 \\
+CMA & 39.71 & 1.4 \\
+F4M & 41.76 & 3.1 \\
+F4M+CMA & \textbf{42.03} & \textbf{4.5} \\
\bottomrule
\end{tabular}
}
\vspace{-6pt}

\label{tab:exp_robust}
\end{table}

%% file: table/exp_dropout.tex
\begin{table}[t]
\centering

\caption{Ablation study on the spatial dropout in CMA and F4M.
}
\vspace{-6pt}

\resizebox{\linewidth}{!}{
\begin{tabular}{l|c|c|c|c|c|c|c}
\hline
Metric & \multicolumn{4}{c|}{$\pi_f$} & \multicolumn{3}{c}{$\pi_c$} \\ \cline{2-8} 
                & 0 & 0.25 & 0.5 & 0.75 & 0 & 0.3  & 0.6 \\ \hline
HICO-DET* & 41.20 & 41.45 & 41.76 & 41.13 & 39.37 & 39.71 & 39.24 \\ \hline
RRM (\%) & 1.9 & 2.6 & \textbf{3.1} & 2.3  & 0.3   & \textbf{1.4} & 0.6 \\ \hline
\end{tabular}
}
\vspace{-6pt}

\label{tab:exp_dropout}
\end{table}

%% file: table/exp_mixup.tex
\begin{table}[t]
\centering

\caption{Comparison between different data augmentation variants in the proposed shift-augmented mixup.}
\vspace{-6pt}

\resizebox{0.95\linewidth}{!}{
\begin{tabular}{c|cc}
\hline
Method & HICO-DET* & RRM (\%) \\
\hline
Baseline & 38.65 & -0.7 \\
Sample synthesis & 36.26 & 0.9 \\
Sample mixing & 39.25 & -0.1 \\
Sample synthesis + Sample mixing & \textbf{39.71} & \textbf{1.4} \\
\hline
\end{tabular}
}
\vspace{-1.5em}

\label{tab:exp_mixup}
\end{table}

%% file: table/exp_domains.tex
\begin{table}[t]
\centering

\caption{Ablation study on the domains in CMA.
}
\vspace{-6pt}

\resizebox{0.7\linewidth}{!}{
\begin{tabular}{c|cc}
\toprule
Number of Domains & HICO-DET* & RRM (\%) \\
\midrule
0 (Baseline) & 38.65 & -0.7 \\
3 & 39.41 & 0.3 \\
6 & 39.53 & 0.8 \\
9 & \textbf{39.82} & 1.0 \\
12 (Default) & 39.71 & \textbf{1.4} \\
\bottomrule
\end{tabular}
}
\vspace{-6pt}

\label{tab:exp_domains}
\end{table}

%% file: table/exp_f4m.tex
\begin{table}[t]
\centering

\caption{Ablation study on the proposed F4M.
}
\vspace{-6pt}

\resizebox{0.9\linewidth}{!}{
  \begin{tabular}{l|cc}
    \toprule
    Method & HICO-DET* & RRM (\%) \\
    \midrule
    Baseline & 38.65 & -0.7 \\
    +Global Token Enhanced Decoder & 40.52 & 2.1 \\
    +Regional Token Enhanced Encoder & 40.15 & 0.9 \\
    +F4M & \textbf{41.76} & \textbf{3.1} \\
    \hline
    +Global Token Enhanced Encoder & 39.14 & -0.1 \\
    +Regional Token Enhanced Decoder & 38.78 & 0.3 \\
    +Vision Foundation Model as Backbone & 35.16 & -0.7 \\
    \bottomrule
  \end{tabular}
}
\vspace{-1.5em}

\label{tab:exp_f4m}
\end{table}

%% file: figure/exp_qualitative.tex
\begin{figure*}
    \centering
    \includegraphics[width=0.65\linewidth]{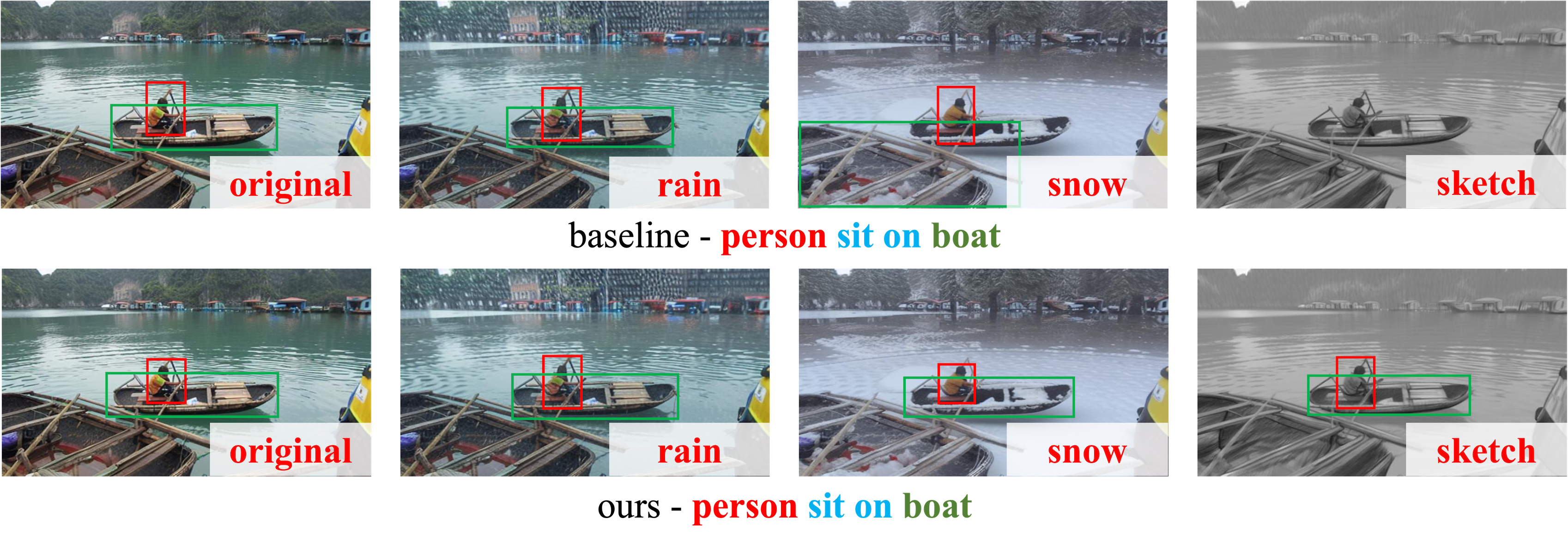}
    \vspace{-1em}
    \caption{
    \textbf{Qualitative comparison between the baseline and the proposed method.}
    The predictions from the baseline are vulnerable to domain changes, while those from the improved model are not.
    Discussed in~\cref{subsec:exp_analysis}.
    More cases in~\supp.
    }
    \vspace{-1em}
    \label{fig:exp_qualitative}
\end{figure*}

%% file: sec/7_conclusion.tex
\section{Conclusion}
\label{sec:conclusion}

This work addresses the research gap concerning the robustness of HOI detection to distribution shift in real-world scenarios. It is the first to highlight the importance of the distribution shift problem for practical applications and conduct relevant studies accordingly. We introduce the first benchmark for assessing such robustness, providing a more realistic and comprehensive evaluation of distribution shifts.
We evaluate the robustness of various existing methods, revealing their limitations and analyzing the characteristics of different frameworks.
Based on the analyses, we propose an approach to boost such robustness, which is plug-and-play and universally applicable to various frameworks. It utilizes existing vision foundation models in a novel way, and synthesizes samples helpful for learning domain-generalizable features.
We verify its effectiveness in both standard performance and robustness through exhaustive experiments.
We discuss the limitations and social impacts of this work in~\supp{} Sec. V.

%% file: sec/X_suppl.tex
\clearpage

\ifCLASSOPTIONcaptionsoff
  \newpage
\fi

\newpage
{\appendices

\section*{Overview}
\begin{itemize}
    \item \textbf{Breakdown over Method Components} (\cref{sec:supp_breakdown}): The complete experiments to analyze how different components of HOI detection methods react to distribution shift.
    \item \textbf{Elaboration on F4M} (\cref{sec:supp_f4m}): The complete information for F4M.
    \item \textbf{Evaluation Setting on Standard HOI Detection} (\cref{sec:supp_setting}): More details regarding the evaluation setting on standard HOI detection.
    \item \textbf{Supplemental Experiments} (\cref{sec:supp_exp}): Additional experiments.
    \item \textbf{Limitations and Social Impact} (\cref{sec:supp_impact}): Discussion on the limitations and potential social impact of this work.

\end{itemize}

\section{Breakdown over Method Components}
\label{sec:supp_breakdown}

\input{table/analysis_pipeline}
\input{table/analysis_backbone}

Here we investigate in detail how the differences in models and data can affect robustness against distribution shift.

\noindent \textbf{Pipelines.}
HOI detectors are primarily categorized by their pipelines.
There are three kinds of pipelines in HOI detection: two-stage~\cite{li2019hoitransferable,zhang2021hoispatially}, one-stage~\cite{kim2020hoiuniondet,liao2020hoippdm}, and end-to-end~\cite{tamura2021hoiqpic,zhuang2023compositional}, as reviewed in \cref{subsec:review_hoi}.
%
Through controlled experiments modifying QPIC~\cite{tamura2021hoiqpic} (see \cref{tab:analysis_pipeline}), we find that pipeline design significantly affects original domain performance but has a limited impact on robustness,
and the two-stage architecture shows marginally better robustness.

\noindent \textbf{Backbones.}
%
Consistent with prior studies~\cite{mao2023robust-coco-o,angarano2024back2bones}, our analysis (\cref{tab:analysis_backbone}) demonstrates that larger, more powerful backbones consistently improve RRM scores, indicating their crucial role in feature extraction for robust HOI detection.

\input{table/analysis_detector}
\input{table/analysis_classifier}

\noindent \textbf{Object detectors.}
The object detectors used in HOI detection methods can be separated into anchor-based~\cite{girshick2015fast,ren2015faster}, anchor-free~\cite{duan2019centernet}, and detection transformer~\cite{carion2020end,zhu2020deformable}.
%
\cref{tab:analysis_detector} reveals detector architecture has minimal impact on HOI robustness, contrasting with object detection findings~\cite{mao2023robust-coco-o}. This highlights unique distribution shift challenges in HOI tasks.

\noindent \textbf{Interaction features.}
Interaction classification can be separated into GNN-based~\cite{zhang2021hoispatially}, transformer-based~\cite{zhang2022hoiupt,zhang2023hoipvic}, and others~\cite{wan2019hoipose,liang2022joint}.
To examine this component with less interference, we adopt two-stage methods which disentangle object detection and interaction classification.
%
\cref{tab:analysis_classifier} shows that the classifier design affects the original domain performance but not the robustness metrics.

\input{table/analysis_data}
\input{table/analysis_vl}

\noindent \textbf{Training data.}
Analysis of additional pre-training data (\cref{tab:analysis_data}) demonstrates such methods~\cite{li2024hoidisentangled} achieve better robustness, suggesting data scaling as a promising direction.

\noindent \textbf{Vision-language knowledge.}
%
Despite CLIP's inherent robustness~\cite{radford2021CLIP,crabbe2024interpreting-clip-robust,tu2024closer-look-clip-robust}, methods incorporating VL knowledge~\cite{liao2022hoigen,ning2023hoiclip,mao2023clip4hoi} show RRM scores near or below average (\cref{tab:analysis_vl}), indicating current limitations in effectively transferring VL robustness to HOI tasks.
This highlights the importance of designing methods that effectively leverage CLIP to boost distribution shift robustness.

\section{Elaboration on F4M}
\label{sec:supp_f4m}

\subsection{Preliminaries}

\noindent \textbf{HOI Detection Transformer.}
Recent HOI detection methods~\cite{tamura2021hoiqpic,liao2022hoigen,xie2023cqlhoi} are almost entirely based on detection Transformers~\cite{carion2020end,zhu2020deformable}. They typically consist of three parts: the backbone network, the encoder, and the decoder. The backbone can be any architecture that generates feature maps; the encoder enhances feature maps using attention mechanisms; and the decoder takes the enhanced feature maps and some object or interaction instance queries as input, refining the instance queries layer by layer and predicting category labels and box coordinates for each query. During training, each instance query is optimized for a single object or background.

\noindent \textbf{Vision Transformer}.
Most VFMs~\cite{radford2021CLIP,fang2023eva,fang2024eva02,sun2023evaclip,oquab2023dinov2,touvron2022deitiii} are based on Vision Transformer (ViT). Unlike CNNs, ViT first partitions images into patches and projects each patch into a token. These tokens are flattened spatially and modeled sequentially with a Transformer encoder. Besides regional tokens, a learnable class token is usually appended at the beginning of the sequence. Regional tokens retain local details of each patch, while the class token represents the overall information of the entire image. We utilize both class and regional tokens in VFMs to enhance HOI detectors.

\subsection{Query Design of Global Token Enhanced Decoder}

Here we elaborate more on the technical design for the global token enhanced decoder in the proposed F4M. Its image query involves 4 types of designs.

Query Type 1: The image is resized to match the input size of the foundation model. Then, in each decoder layer, the image query is projected into the same dimension as the instance queries, and these two types of queries are concatenated before being fed into the self-attention module. In the subsequent self-attention, instance queries can adaptively interact with the image query to extract high-level image understanding information from the foundation model. Finally, the image query is discarded after passing through the self-attention module.

Query Type 2: To further preserve fine-grained details in the global token, we follow previous works~\cite{fu2024frozendetr,chen2024internvl2} utilizing CLIP, to divide the image uniformly into multiple sub-images and assign an image query to each sub-image. For example, if the image is divided into 2×2 sub-images, we obtain five image queries, including one original global image query and four local image queries.

Query Type 3 and 4: Repeated inference of the VFM in Query Type 2 is time-consuming, as they are usually quite large. We adopt two faster implementations to overcome this limitation. In the first one, the regional tokens are spatially split into multiple groups corresponding to the sub-images, where the average features of each token group serve as its local image query. This implementation is called Query Type 3.
The second one uses multiple global tokens. Masked attention operations are applied to ensure each global token focuses on its corresponding sub-image by restricting interactions to the local area. Both fast implementations can acquire global and all local image queries within a single inference process, significantly reducing computational costs. This implementation is called Query Type 4.

By default, we use Query Type 4, which is a good balance between fine-grained details and lightweight inference costs. We verify this in the following experiments.

\section{Evaluation Setting on Standard HOI Detection}
\label{sec:supp_setting}

Following most previous works~\cite{tamura2021hoiqpic,xie2023cqlhoi,liao2022hoigen}, we adopt the widely-used HICO-DET~\cite{chao2018hoihicodet} and V-COCO~\cite{gupta2015hoivcoco} separately, both containing a training set and a test set.
HICO-DET contains 47,776 images, with 38,118 for training and 9,658 for testing. There are 600 HOI categories in HICO-DET, consisting of 117 interaction classes and 80 object classes. Each HOI class is composed of an interaction and an object.
V-COCO is a subset of MS-COCO~\cite{lin2014microsoft} with HOI annotations, including 10,346 images (2,533 for training, 2,867 for validation, and 4,946 for testing). It has 80 object categories same as HICO-DET, and 29 interaction categories.

For HICO-DET, we adopt the commonly used mAP metric. Each prediction is a \textlangle{}human, interaction, object\textrangle{} triplet. A prediction is a true positive only when the human and object bounding boxes both have IoU \textgreater 0.5 w.r.t. ground truth and the interaction classification result is correct. We evaluate the performance in two different settings following~\cite{chao2018hoihicodet}. In the \textit{known object} setting, for each HOI category, we evaluate the prediction only on the images containing the target object category. In \textit{default} setting, the detection result of each category is evaluated on the full test set. In each setting, we report the mAP over (1) all 600 HOI categories (Full), (2) 138 categories with less than 10 training samples (Rare), and (3) the remaining 462 categories (Non-rare).
For V-COCO, we use the role mAP following~\cite{gupta2015hoivcoco}, under both scenario \#1 (including objects) and \#2 (ignoring objects). The performance is evaluated using its official evaluation toolkit.

\section{Supplemental Experiments}
\label{sec:supp_exp}

\input{table/exp_finetuning}

\input{figure/analysis_rrm}
\subsection{Performance on Robust-HICO and HICO-DET*}
\cref{fig:analysis_rrm} reveals that despite significant progress on HICO-DET, modern methods show no measurable improvement in distribution shift robustness, as evidenced by their RRM values.

\subsection{More Ablation Studies}

We perform more ablation studies in \supp, including the influence of different VFMs, the importance of freezing the VFM, and the effectiveness of the query design in our global token enhanced decoder.

\input{table/exp_foundation}
\noindent \textbf{Different VFMs.}
Previous works primarily utilize only CLIP among various VFMs.
However, there are many types of VFMs, trained with different tasks and data, so it is important to explore which would be useful for such robustness.
Six representative foundation models with different pre-training methods were chosen: CLIP~\cite{radford2021CLIP} and EVA-CLIP~\cite{sun2023evaclip} (image-text contrastive learning); DEiT-III~\cite{touvron2022deitiii} (supervised image classification pre-training); MAE~\cite{he2022masked} and DINOv2~\cite{oquab2023dinov2} (masked data modeling); SAM~\cite{Kirillov2023SAM} (promptable segmentation).
For models that do not use global tokens during pre-training (such as MAE), average regional tokens are used as image queries.

According to~\cref{tab:exp_foundation}, we found VFMs pretrained with labels, such as DEiT-III, CLIP, and EVA-CLIP, perform better than self-supervised models like DINOv2. This might be because self-supervised models lack high-level semantics from human supervision, making these features difficult to apply to downstream tasks without fine-tuning. Another possible reason is that masked data modeling pre-trained models may focus more on local texture details, resulting in less robustness to distribution shifts compared to multimodal contrastive learning models like CLIP. Additionally, CLIP and EVA-CLIP perform similarly and both outperform DEiT-III significantly, possibly due to their large-scale image-text contrastive learning pre-training. So we use CLIP by default for ablation.

\noindent \textbf{Combining multiple VFMs.}
We conduct more investigations to see if combining multiple foundation models could further improve performance, as VFMs trained on different tasks might understand images differently and extract features complementary to each other.
As shown in~\cref{tab:exp_foundation}, using more than one VFM indeed enhances the performance and distribution shift robustness. This suggests potential complementarity among different foundation models in understanding images. Additionally, combining the relatively effective models CLIP and EVA-CLIP yields less than ideal results, possibly due to their similar training tasks providing overlapping robustness features; the best combination effect comes from pairing the image-text contrastive learning CLIP with the self-supervised learning DINOv2, likely because their distinct training methods produce complementary features. Due to increased costs, we do not explore three or more VFMs.

\noindent \textbf{Finetuning or freezing VFMs.}
As shown in~\cref{tab:exp_finetuning}, fine-tuning the foundation model results in worse performance than freezing it. One possible reason is that training with HOI detection data may disrupt the pre-trained representations in the foundation model; another possible explanation is that fine-tuning on a single domain affects the robust features of the VFM itself for generalizing across different domains.

\input{table/exp_query_type}
\input{table/exp_input_size}
\input{figure/exp_qualitative_more}
\input{figure/exp_case}

\noindent \textbf{Different image query.}
As discussed in Sec. V-C of the main paper, using local image queries for sub-images can preserve more fine-grained details. \cref{tab:exp_query_type} shows the results of different implementations. Among them, query type 2, which crops sub-images from the original image, significantly improves the effect compared to query type 1, but it noticeably impacts training and inference speed since the image must pass through the entire foundation model multiple times. In contrast, using average regional tokens (query type 3) or multiple global tokens (query type 4) saves considerable time as only one pass is required, though the former is less effective than the latter. Therefore, in the following experiments, query type 4, which uses multiple global tokens to provide multiple image queries, is used by default.

\noindent \textbf{Input size for VFMs}.
We also explore the input image size of the foundation model through ablation studies. In Table~\cref{tab:exp_input_size}, for F4M using only CLIP, the input sizes are set to 224, 336, 448, and 484. It was found that larger input image sizes are not necessarily better: as the size increases, performance on the original domain and the robustness metric RRM first rise and then fall. On one hand, since the foundation model is pre-trained at lower resolutions, the image size should not be too large. On the other hand, the foundation model provides high-level, domain-low-related image semantic understanding rather than domain-high-related textures and other details. Thus, very large input sizes are unnecessary. In addition, larger input sizes bring substantial computational burdens. By default, we set the input image size to 336.

\noindent \textbf{More qualitative cases.}
In Sec. VI-C of the main paper, we present a pair of qualitative cases, to show the proposed model helps produce correct and consistent instances on different domains. We show more cases in~\cref{fig:exp_qualitative_more}.

\noindent \textbf{Error types of specific cases.}
In~\cref{fig:exp_case}, we compare the variants of the proposed method on two different domains, the original one and a shifted one (watercolor).
In the first example, for the original domain, the interaction classification error of the baseline is less than 40\% of the total false predictions. F4M, CMA, and their combination, do not show significant targeted improvements for this type of error (although they are indeed effective in terms of standard performance in the original domain, as verified before).
In contrast, in the watercolor domain, where the proportion of interaction classification errors is significantly higher (over 50\%), these methods demonstrate a clear improvement in reducing the proportion of interaction classification errors.
Among them, F4M significantly reduces the proportion of this type of error; CMA also reduces such errors, though less pronounced than F4M; the combination of the two brings even more significant results, likely because they enhance robustness from complementary perspectives, the features and training samples. In the second example, similar phenomena can be observed: the F4M and CMA methods have a much more prominent effect in reducing interaction classification errors in the watercolor domain than in the original domain.
Overall, these cases show the proposed method improves robustness mainly by reducing the interaction classification errors on the shifted domains.

\section{Limitations and Social Impact}
\label{sec:supp_impact}

\noindent \textbf{Limitations.}
There are also some limitations in this work.
Firstly, despite our efforts, it is difficult to simulate all possible types of distribution shift during HOI detection applications.
Secondly, the proposed approach universally boosts the robustness of existing models, but the strongest robustness we can obtain is still limited.
These are impossible to finish in one work. We would like to continuously investigate the robustness of HOI detection in the future, and we hope this work will inspire more research in this direction.

\noindent \textbf{Social Impact.}
The proposed method has no direct potential negative impact.
However, we note that it focuses on HOI detection in practical scenarios, which is related to the recognition and detection of objects and actions in real life. Thus, the technology involved may be subjected to some potential abuse in applications like video surveillance.

}


%% file: table/analysis_pipeline.tex
\begin{table}[t]
\centering

\caption{Comparison of several methods under the influence of different HOI detection pipelines.
}
\vspace{-6pt}

\resizebox{\linewidth}{!}{
  \begin{tabular}{c|c|ccc}
    \toprule
    Method & Pipeline & HICO-DET* & Robust-HICO & RRM (\%) \\
    \midrule
    QPIC~\cite{tamura2021hoiqpic} & end-to-end & 33.64 & 22.29 & -1.7 \\
    QPIC** & two-stage & 32.75 & 21.93 & \textbf{-1.0} \\
    QPIC* & one-stage & 30.49 & 21.34 & -1.9 \\
    \bottomrule
  \end{tabular}
}
\vspace{-6pt}

\label{tab:analysis_pipeline}
\end{table}

%% file: table/analysis_backbone.tex
\begin{table}[t]
\centering

\caption{Comparison of several methods under the influence of different backbone networks.
}
\vspace{-6pt}

\resizebox{\linewidth}{!}{
  \begin{tabular}{c|c|ccc}
    \toprule
    Method & Backbone & HICO-DET* & Robust-HICO & RRM (\%) \\
    \midrule
    \multirow{3}{*}{SCG~\cite{zhang2021hoispatially}} & ResNet50 & 30.12 & 20.56 & 0.3 \\
                         & ResNet101 & 31.23 & 21.57 & 1.1 \\
                         & Swin-L & 36.08 & 25.62 & \textbf{3.0} \\
    \hline
    \multirow{3}{*}{QPIC~\cite{tamura2021hoiqpic}} & ResNet50 & 33.64 & 22.29 & -1.7 \\
                          & ResNet101 & 34.20 & 23.38 & 0.4 \\
                          & Swin-L & 37.68 & 26.35 & \textbf{1.9} \\
    \hline
    \multirow{3}{*}{PViC~\cite{zhang2023hoipvic}} & ResNet50 & 38.65 & 26.02 & -0.7 \\
                          & ResNet101 & 39.84 & 27.64 & 1.4 \\
                          & Swin-L & 45.37 & 32.81 & \textbf{4.3} \\
    \bottomrule
  \end{tabular}
}
\vspace{-6pt}

\label{tab:analysis_backbone}
\end{table}

%% file: table/analysis_detector.tex
\begin{table}[t]
\centering

\caption{Comparison of several methods under the influence of different object detectors.}
\vspace{-6pt}

\resizebox{\linewidth}{!}{
\begin{tabular}{c|c|ccc}
    \toprule
    Method & Detector & HICO-DET* & Robust-HICO & RRM (\%) \\
    \midrule
    \multirow{3}{*}{SCG~\cite{zhang2021hoispatially}} & Faster R-CNN & 30.12 & 20.56 & \textbf{0.3} \\
                 & CenterNet & 29.24 & 19.85 & -0.1 \\
                 & DETR & 29.45 & 20.05 & 0.1 \\
    \midrule
    \multirow{3}{*}{ViPLO~\cite{park2023hoiviplo}} & Faster R-CNN & 36.07 & 24.95 & \textbf{1.2} \\
                   & CenterNet & 35.26 & 24.25 & 0.8 \\
                   & DETR & 35.71 & 24.54 & 0.7 \\
    \bottomrule
  \end{tabular}
}
\vspace{-6pt}

\label{tab:analysis_detector}
\end{table}

%% file: table/analysis_classifier.tex
\begin{table}[t]
\centering

\caption{Comparison of different interaction classifier for two-stage HOI detectors. DETR~\cite{carion2020end} is adopted as the first stage network (including the backbone) for fair alignment.
}
\vspace{-6pt}

\resizebox{0.9\linewidth}{!}{
  \begin{tabular}{c|ccc}
    \toprule
    Interaction Classifier & HICO-DET* & Robust-HICO & RRM (\%) \\
    \midrule
    SCG~\cite{zhang2021hoispatially} & 31.46 & 20.24 & \textbf{-0.3} \\
    UPT~\cite{zhang2022hoiupt} & 34.89 & 23.51 & -0.6 \\
    PViC~\cite{zhang2023hoipvic} & 38.65 & 26.02 & -0.7 \\
    \bottomrule
  \end{tabular}
}
\vspace{-6pt}

\label{tab:analysis_classifier}
\end{table}

%% file: table/analysis_data.tex
\begin{table}[t]
  \centering

  \caption{Effect of more HOI detection training data.}
  \vspace{-6pt}

  \resizebox{\linewidth}{!}{
  \begin{tabular}{c|c|ccc}
    \toprule
    Method & Extra Data & HICO-DET* & Robust-HICO & RRM (\%) \\
    \midrule
    CDN~\cite{zhang2021hoimining} & - & 35.32 & 23.29 & -2.1 \\
    CDN~\cite{zhang2021hoimining} & DP-HOI (484k) & 37.84 & 25.91 & \textbf{0.5} \\
    \hline
    HOICLIP~\cite{ning2023hoiclip} & - & 39.62 & 26.35 & -1.5 \\
    HOICLIP~\cite{ning2023hoiclip} & DP-HOI (484k) & 41.12 & 28.72 & \textbf{1.8} \\
    \bottomrule
  \end{tabular}
  }
\vspace{-6pt}

\label{tab:analysis_data}
\end{table}

%% file: table/analysis_vl.tex
\begin{table}[t]
\centering

\caption{Comparison of different HOI detection methods utilizing VL knowledge in CLIP~\cite{radford2021CLIP}.
}
\vspace{-6pt}

\resizebox{0.9\linewidth}{!}{
\begin{tabular}{c|ccc}
\toprule
Method & HICO-DET* & Robust-HICO & RRM (\%) \\
\midrule
GEN-VLKT~\cite{liao2022hoigen} & 38.50 & 25.63 & -1.4 \\
HOICLIP~\cite{ning2023hoiclip} & 39.62 & 26.35 & -1.5 \\
LogicHOI~\cite{li2023hoineurallogic} & 40.07 & 26.37 & -2.2 \\
CLIP4HOI~\cite{mao2023clip4hoi} & 39.87 & 27.20 & \textbf{0.2} \\
\bottomrule
\end{tabular}
}
\vspace{-6pt}

\label{tab:analysis_vl}
\end{table}

%% file: table/exp_finetuning.tex
\begin{table}[t]
\centering

\caption{Ablation study on whether finetuning the vision foundation model in F4M or not.
}

\resizebox{0.8\linewidth}{!}{
\begin{tabular}{l|c|cc}
\toprule
Method & Fine-Tuning VFM & HICO-DET* & RRM (\%) \\
\midrule
PViC~\cite{zhang2023hoipvic} & - & 38.65 & -0.7 \\
\hline
+F4M & Yes & 40.23 & -0.4 \\
& No & \textbf{41.76} & \textbf{3.1} \\
\bottomrule
\end{tabular}
}
\vspace{-6pt}

\label{tab:exp_finetuning}
\end{table}

%% file: figure/analysis_rrm.tex
\begin{figure*}
    \centering
    \includegraphics[width=0.9\linewidth]{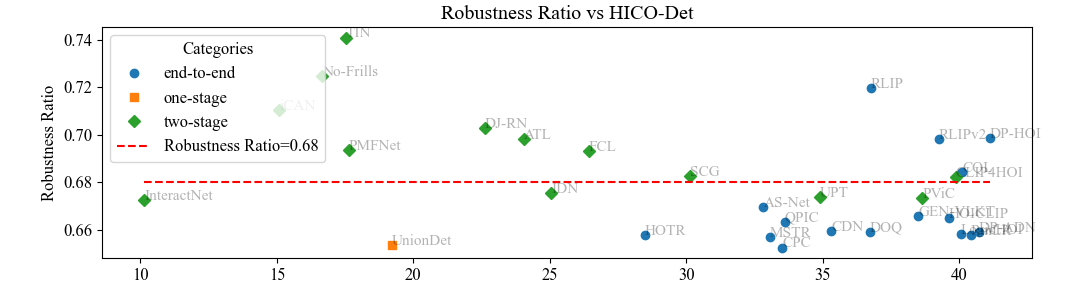}
    \caption{
    \textbf{Evaluation results of different methods on Robust-HICO and HICO-DET*.}
    To show the development of recent methods regarding distribution shift robustness, we only visualize the models with ResNet50~\cite{he2016deep} backbones.
    Discussed in Sec. IV-A of the paper.
    }
    \vspace{-6pt}
    \label{fig:analysis_rrm}
\end{figure*}

%% file: table/exp_foundation.tex
\begin{table}[t]
\centering

\caption{Ablation study on the foundation models used in F4M.
}
\vspace{-6pt}

\resizebox{0.8\linewidth}{!}{
\begin{tabular}{l|cc}
\toprule
Vision Foundation Model & HICO-DET* & RRM (\%) \\
\midrule
- & 38.65 & -0.7 \\
CLIP~\cite{radford2021CLIP} & 41.04 & \textbf{2.3} \\
EVA-CLIP~\cite{fang2024eva02} & \textbf{41.12} & 2.0 \\
DEiT-III~\cite{touvron2022deitiii} & 40.75 & 1.6 \\
MAE~\cite{he2022masked} & 39.78 & 0.8 \\
DINOv2~\cite{oquab2023dinov2} & 40.11 & 1.1 \\
\hline
CLIP~\cite{radford2021CLIP} + EVA-CLIP~\cite{fang2024eva02} & 41.42 & 2.5 \\
CLIP~\cite{radford2021CLIP} + DINOv2~\cite{oquab2023dinov2} & \textbf{41.76} & \textbf{3.1} \\
CLIP~\cite{radford2021CLIP} + DEIT-III~\cite{touvron2022deitiii} & 41.52 & 2.4 \\
DINOv2~\cite{oquab2023dinov2} + DEIT-III~\cite{touvron2022deitiii} & 40.89 & 1.9 \\
\bottomrule
\end{tabular}
}
\vspace{-6pt}

\label{tab:exp_foundation}
\end{table}

%% file: table/exp_query_type.tex
\begin{table}[t]
\centering

\caption{Ablation study on the image query type in F4M. \# Inferences denotes the number of inferences per sample.
}

\resizebox{0.8\linewidth}{!}{
\begin{tabular}{l|ccc}
\toprule
Image Query & HICO-DET* & RRM (\%) & \#Inferences \\
\midrule
Query Type 1 & 40.54 & 1.4 & 1 \\
Query Type 2 & 42.01 & 3.0 & 1+4 \\
Query Type 3 & 41.21 & 1.9 & 1 \\
Query Type 4 & 41.76 & 3.1 & 1 \\
\bottomrule
\end{tabular}
}
\vspace{-6pt}

\label{tab:exp_query_type}
\end{table}

%% file: table/exp_input_size.tex
\begin{table}[t]
\centering

\caption{Ablation study on the input size in F4M.
}

\resizebox{0.6\linewidth}{!}{
\begin{tabular}{c|cc}
\toprule
Input Size & HICO-DET* & RRM (\%) \\
\midrule
224 & 41.15 & 2.3 \\
336 & 41.76 & \textbf{3.1} \\
448 & \textbf{41.91} & 2.9 \\
484 & 41.56 & 2.6 \\
\bottomrule
\end{tabular}
}
\vspace{-6pt}

\label{tab:exp_input_size}
\end{table}

%% file: figure/exp_qualitative_more.tex
\begin{figure*}
    \centering
    \includegraphics[width=0.8\linewidth]{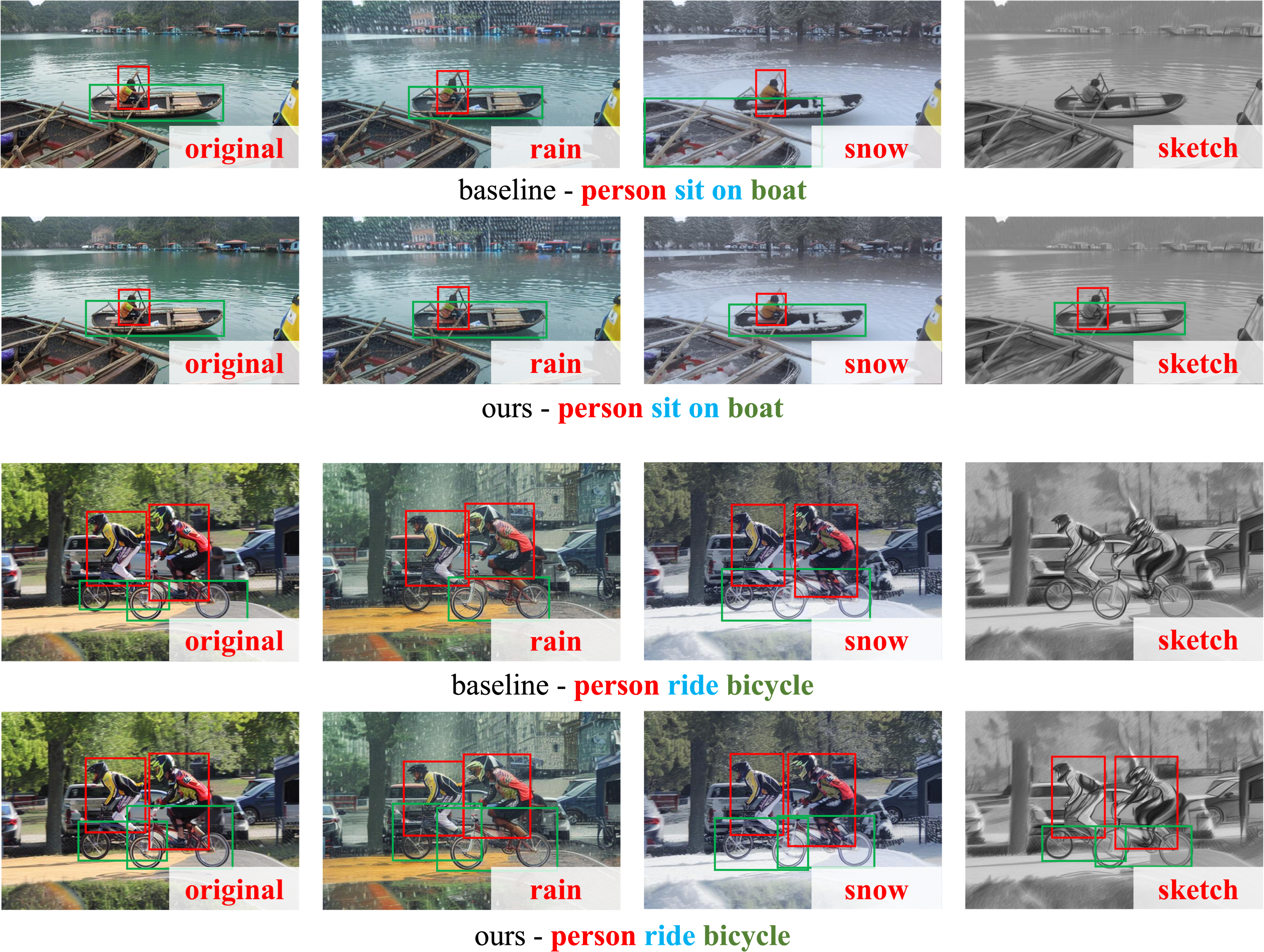}
    \caption{
    \textbf{Qualitative comparison between the baseline and the proposed method.}
    The first two rows show an example of ``person sit on boat'' while the latter two rows show ``person ride bicycle''.
    }
    \vspace{-6pt}
    \label{fig:exp_qualitative_more}
\end{figure*}

%% file: figure/exp_case.tex
\begin{figure*}
    \centering
    \includegraphics[width=0.75\linewidth]{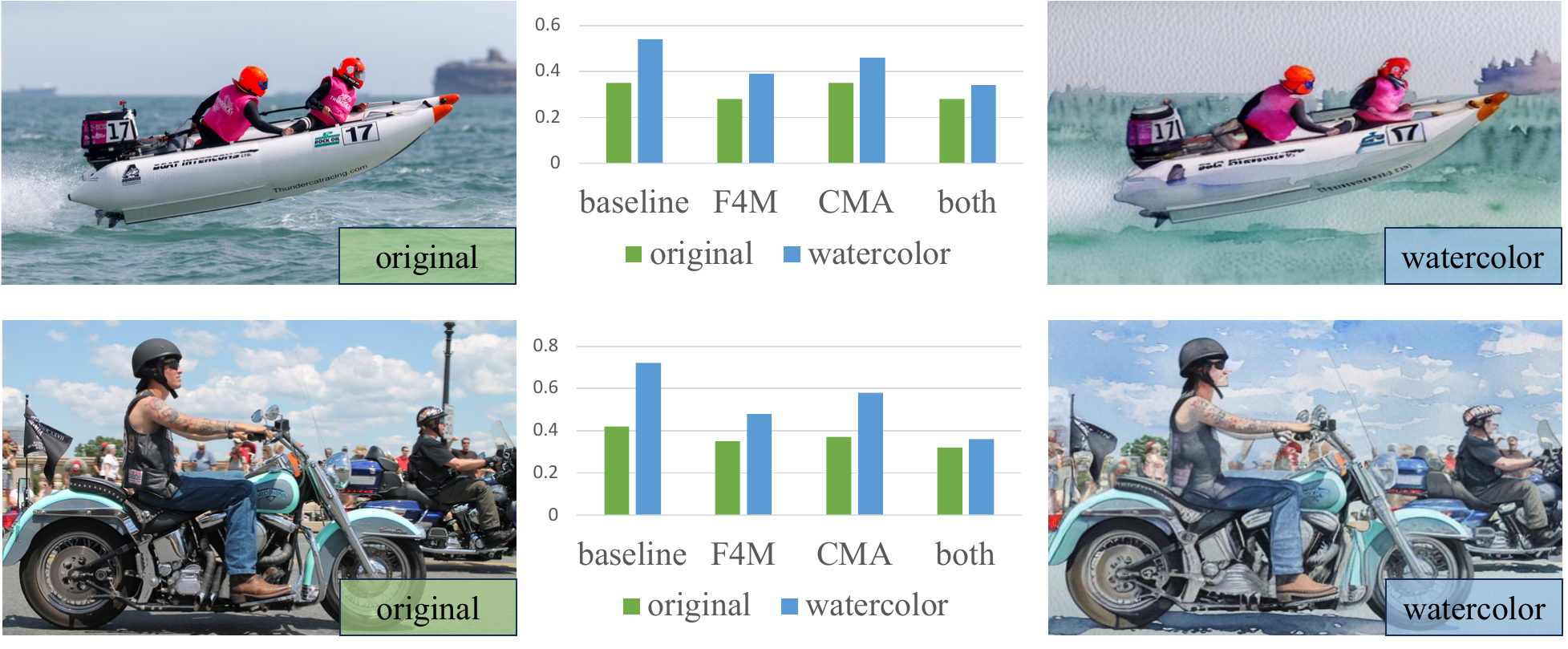}
    \caption{
    \textbf{Qualitative comparison between different parts of the proposed method.}
    The comparison is over the original domain and the shifted watercolor domain in two specific cases.
    }
    \vspace{-6pt}
    \label{fig:exp_case}
\end{figure*}